%% file: main.tex
\documentclass{article} % For LaTeX2e
\usepackage{iclr2022_conference,times}

\input{math_commands.tex}

\usepackage[pagebackref=true,colorlinks,bookmarks=false,citecolor=blue,linkcolor=blue]{hyperref}
\usepackage{paralist}
\usepackage{booktabs}
\usepackage{xspace} % for \eg, \ie, etc.
\usepackage{stmaryrd}
\usepackage[normalem]{ulem}
\usepackage[tableposition=top]{caption}

\usepackage[title,titletoc,page]{appendix}
\usepackage{tocloft}

\usepackage{multirow}
\usepackage{makecell}
\usepackage{soul}
\usepackage{graphicx}
\usepackage{bbold}
\usepackage[bottom]{footmisc} % Keep footnote at bottom
\usepackage{wrapfig}
\usepackage{enumitem}

\usepackage[frozencache,cachedir=.]{minted} % finalizecache
\input{macro}

\title{\modelName: Towards Infinite-Pixel\\Image Synthesis}

\author{%
Chieh Hubert Lin$^{1}$ \,
Hsin-Ying Lee$^{2}$ \,
Yen-Chi Cheng$^{3}$ \,
Sergey Tulyakov$^{2}$ \,
Ming-Hsuan Yang$^{1,4,5}$
\\
$^{1}$UC Merced \,
$^{2}$Snap Inc. \,
$^{3}$Carnegie Mellon University \,
$^{4}$Yonsei University \,
$^{5}$Google Research \,
}

\iclrfinalcopy % Uncomment for camera-ready version, but NOT for submission.
\begin{document}

\maketitle
\input{0-abstract}

%%%%%%%%%%%%%% Teaser
{
    \vspace{-3mm} 
    \centering
    \includegraphics[width=.98\linewidth]{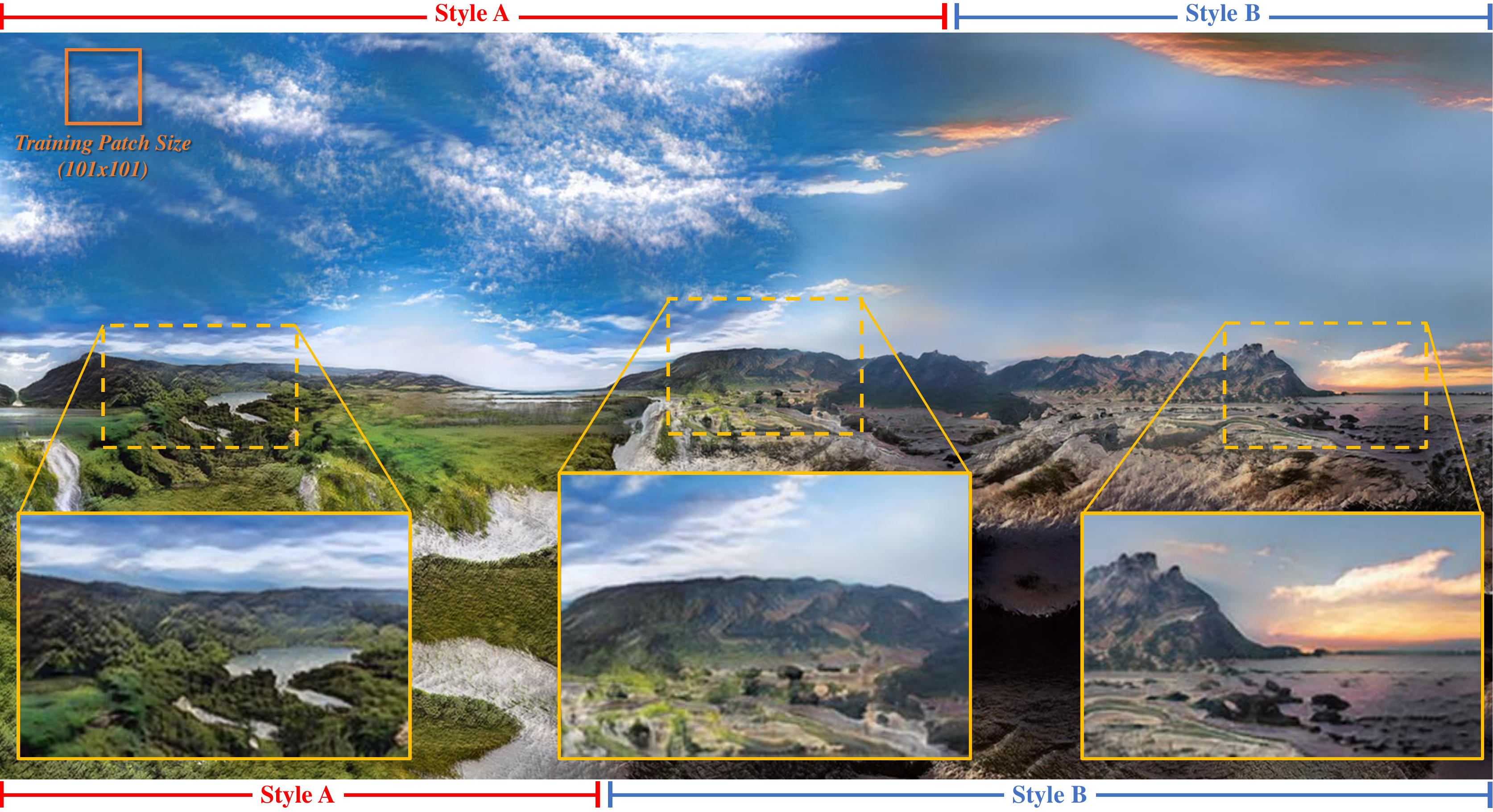}
    \vspace{-1em}
    \captionof{figure}{
        \textbf{Synthesizing infinite-pixel images from finite-sized training data.}
        A 1024$\times$2048 image composed of 242 patches, independently synthesized by InfinityGAN with spatial fusion of two styles. 
        The generator is trained on 101$\times$101 patches (\eg marked in {\color{orange} top-left}) sampled from 197$\times$197 real images.
        Note that training and inference (of any size) are performed on a single GTX TITAN X GPU. Zoom-in for better experience.
        \label{fig:teaser}
    }
    \vspace{-2mm}
}

\addtocontents{toc}{\protect\setcounter{tocdepth}{0}} % Disable toc 
\input{1-introduction}
\input{2-related_work}
\input{3-method}
\input{4-experiment}
\input{5-conclusion}

\section{Acknowledgements}
This work is supported in part by the NSF CAREER Grant \#1149783 and a gift from Snap Inc.
\input{7-ethics-statement}

% \bibliography{a-citation}
% \bibliographystyle{iclr2022_conference}
\bibliographystyle{iclr2022_arxiv}

\input{main.bbl}
\clearpage

ArXiv latex compiler does not accept our supplementary.

Please find the full version of the paper on OpenReview with a better experience: \href{https://openreview.net/forum?id=ufGMqIM0a4b }{https://openreview.net/forum?id=ufGMqIM0a4b }.

% \clearpage
% \appendix

% % Change subsection format
% % \addcontentsline{toc}{section}{Appendices}
% % \renewcommand{\thesubsection}{\Alph{subsection}}
% \renewcommand*\cftsecnumwidth{18pt} % default ~15
% \renewcommand{\cftsecleader}{\cftdotfill{\cftsecdotsep}}
% \renewcommand\cftsecdotsep{\cftdot}
% % \renewcommand\cftsubsecdotsep{\cftdot}
% \renewcommand{\contentsname}{Appendix Table of Contents}
% % \addto\captionsenglish{\def\contentsname{Appendix Table of Contents}} %! Needed for babel? https://tex.stackexchange.com/questions/35903/formatting-the-title-of-the-toc

% % Enable ToC
% \addtocontents{toc}{\protect\setcounter{tocdepth}{1}}
% \tableofcontents

% % % Commented to speed up compilation
% \newpage
% \input{6-supp}

\end{document}

%% file: math_commands.tex
\usepackage{amsmath,amsfonts,bm}

\def\1{\bm{1}}

\DeclareMathAlphabet{\mathsfit}{\encodingdefault}{\sfdefault}{m}{sl}
\SetMathAlphabet{\mathsfit}{bold}{\encodingdefault}{\sfdefault}{bx}{n}

%% file: macro.tex
\newcommand{\loss}[1]{\mathcal{L}_\mathrm{#1}}
\newcommand{\lambdaw}[1]{\lambda_\mathrm{#1}}
\newcommand{\structuresyn}{G_\mathrm{S}}
\newcommand{\texturesyn}{G_\mathrm{T}}
\newcommand{\globalz}{\mathbf{z}_\mathrm{g}}
\newcommand{\localz}{\mathbf{z}_\mathrm{l}}
\newcommand{\localzfir}{\mathbf{z}_{\mathrm{l}_1}}
\newcommand{\localzsec}{\mathbf{z}_{\mathrm{l}_2}}
\newcommand{\noisez}{\mathbf{z}_\mathrm{n}}
\newcommand{\structurez}{\mathbf{z}_\mathrm{S}}

\newcommand{\coordgrid}{\mathbf{c}}
\newcommand{\patch}{\mathbf{p}_\mathrm{c}}

\newcommand{\inout}{In\&Out}

\usepackage{xcolor}
\usepackage{array}
\usepackage{xspace}

\newcommand\blfootnote[1]{%
  \begingroup
  \renewcommand\thefootnote{}\footnote{#1}%
  \addtocounter{footnote}{-1}%
  \endgroup
}

\newcommand{\modelName}{InfinityGAN\xspace}

\def\signed #1{{\leavevmode\unskip\nobreak\hfil\penalty50\hskip2em
  \hbox{}\nobreak\hfil -- #1%
  \parfillskip=0pt \finalhyphendemerits=0 \endgraf}}
\newsavebox\mybox
\newenvironment{signedquote}[1]
  {\savebox\mybox{#1}\begin{quote}}
  {\signed{\usebox\mybox}\end{quote}}

\definecolor{ImproveGreen}{rgb}{0.2235, 0.7094, 0.2902}

\usepackage{color}
\definecolor{crimson}{rgb}{0.86, 0.08, 0.24}
\definecolor{gray}{rgb}{0.5,0.5,0.5}
\definecolor{green}{rgb}{0, 0.4, 0}
\definecolor{orange}{rgb}{1, 0.5, 0}
\definecolor{mahogany}{rgb}{0.75, 0.25, 0.0}
\definecolor{purple}{rgb}{0.6, 0, 0.6}
\definecolor{darkgreen}{rgb}{0, 0.4, 0}
\definecolor{frenchblue}{rgb}{0.0, 0.45, 0.73}
\definecolor{red}{rgb}{1,0,0}
\definecolor{yellow}{rgb}{1,1,0}
\definecolor{magenta}{rgb}{1,0,1}
\definecolor{pink}{rgb}{1,0.412,0.706}

\newcommand{\mycomment}[1]{}
\newcommand{\comment}[1]{}

\makeatletter
\DeclareRobustCommand\onedot{\futurelet\@let@token\@onedot}
\def\@onedot{\ifx\@let@token.\else.\null\fi\xspace}

\def\eg{\emph{e.g}\onedot} 
\def\ie{\emph{i.e}\onedot}

\makeatother

\def\eg{e.g.,~}               % for example
\def\ie{i.e.,~}               % that is, in other words
\newlength\paramargin
\newlength\figmargin
\newlength\subfigmargin
\newlength\secmargin
\newlength\subsecmargin
\newlength\tabmargin
\newlength\eqmargin

\newcommand{\figref}[1]{Figure~\ref{fig:#1}}

\long\def\ignorethis#1{}
\definecolor{crimson}{rgb}{0.86, 0.08, 0.24}
\definecolor{green}{rgb}{0, 0.5, 0.25}
\definecolor{purple}{rgb}{0.75, 0, 1}
\definecolor{orange}{rgb}{1, 0.5, 0.25}
\definecolor{yellow}{rgb}{1, 1, 0}
\definecolor{new_blue}{rgb}{0, 0.5, 1}

%% file: 0-abstract.tex
\begin{abstract}
 \vspace{-3mm}
    We present a novel framework,  InfinityGAN, for arbitrary-sized image generation.
    The task is associated with several key challenges. 
    First, scaling existing models to an arbitrarily large image size is resource-constrained, in terms of both computation and availability of large-field-of-view training data. 
    InfinityGAN trains and infers in a seamless patch-by-patch manner with low computational resources.
    Second, large images should be locally and globally consistent, avoid repetitive patterns, and look realistic. To address these, InfinityGAN disentangles global appearances, local structures, and textures. With this formulation, we can generate images with spatial size and level of details not attainable before. 
    Experimental evaluation validates that InfinityGAN generates images with superior realism compared to baselines and features parallelizable inference.
    Finally, we show several applications unlocked by our approach, such as spatial style fusion, multi-modal outpainting, and image inbetweening.
    All applications can be operated with arbitrary input and output sizes.
    \ificlrfinal
        \blfootnote{
        \scriptsize All codes, datasets, and trained models are publicly available.
        Project page: \href{https://hubert0527.github.io/infinityGAN/ }{https://hubert0527.github.io/infinityGAN/ }}
    \else
        \blfootnote{All codes, datasets, and trained models will be made publicly available.
        }
    \fi
\end{abstract}

\comment{
   We aim to learn a latent generative model that can synthesize arbitrary-sized images.%, and generalize to an infinite resolution.
   %
%   We first investigate several recent studies that are potentially applicable to the task, and analyze their shortcomings. 
   %
   We propose a novel InfinityGAN model which consists of a structure synthesizer inspired by neural implicit functions and a fully-convolutional texture synthesizer.
   We train an InfinityGAN model based on patch supervision at a small size that requires low computational resources, and inference with a constant amount of memory in a generation-by-part manner.
   Extensive qualitative and quantitative results show the InfinityGAN model synthesizes images with more favorable global structures than the other alternatives.
   We then demonstrate that the InfinityGAN model learns a strong image prior that help enhances the performance of image outpainting by a large margin.
   %
   %MH: inference speed-up is okay, but can be further improved... will come to this later...
   Finally, we present applications of InfinityGAN for spatial style fusioning, multi-modal outpainting, image inbetweening, and inference speed-up.
    %   As the training images are of a limited resolution, we accordingly design a novel architecture that disentangles the structure and texture learning, such that part of the model is dedicated in unsupervised learning the structural relationships of objects from images, and 
    
}

%% file: 1-introduction.tex
% \vspace{\secmargin}
\section{Introduction}
\label{sec:intro}
\vspace{\secmargin}

\begin{signedquote}{\textit{Buzz Lightyear}}
    \textit{"To infinity and beyond!"}
    \vspace{-0.5em}
\end{signedquote}

Generative models witness substantial improvements in resolution and level of details. Most improvements come at a price of increased training time~\citep{wgangp,mescheder2018r1reg}, larger model size~\citep{balaji2021understanding}, and stricter data requirements~\citep{karras2018progressive}.
The most recent works synthesize images at 1024$\times$1024 resolution featuring a high level of details and fidelity.
%However, straightforward scaling of such models to even larger sizes is problematic.
However, models generating high resolution images usually still synthesize images of limited field-of-view bounded by the training data.
It is not straightforward to scale these models to generate images of arbitrarily large field-of-view. 
% To provide a remedy and to synthesize an arbitrarily-shaped image, \hubert{several approaches~\citep{shaham2019singan,shocher2019ingan,lin2019cocogan} learn an internal patch distribution by training the generator with image patches rather than the full images.}
% several approaches learn an internal patch distribution of a single image to generate images patch-by-patch. 
Synthesizing infinite-pixel images is constrained by the finite nature of resources. Finite computational resources (\eg memory and training time) set bounds for input receptive field and output size. A further limitation is that there exists no infinite-pixel image dataset.
%\hubert{\st{Collecting a representative dataset with \emph{large} images is a challenge itself. }}
Thus, to generate infinite-pixel images, a model should learn the implicit global structure without direct supervision and under limited computational resources. 

Repetitive texture synthesis methods~\citep{efros1999texture,xian2018texturegan} generalize to large spatial sizes. Yet, such methods are not able to synthesize real-world images. Recent works, such as SinGAN~\citep{shaham2019singan} and InGAN~\citep{shocher2019ingan}, learn an internal patch distribution for image synthesis. 
Although these models can generate images with arbitrary shapes, 
% in a nutshell, both approaches reach visual diversity by permuting and randomizing the memorized repetitive texture. 
%both approaches memorize objects and patterns of the scene with position encoding, then reach visual diversity by permuting and randomizing the repetitive texture. 
in Section~\ref{exp:qualitative}, we show that they do not infer structural relationships well, and fail to construct plausible holistic views with spatially extended latent space. A different approach, COCO-GAN~\citep{lin2019cocogan}, learns a coordinate-conditioned patch distribution for image synthesis. As shown in Figure~\ref{fig:exp-qualitative}, despite the ability to slightly extend images beyond the learned boundary, it fails to maintain the global coherence of the generated images when scaling to a 2$\times$ larger generation size.
% the method is not able to maintain consistency when large (how large?) resolutions are concerned.

\emph{How to generate infinite-pixel images?} Humans are able to guess the whole scene given a partial observation of it. 
%In this case the scene will be holistically coherent, while local colors and details across patches may vary.
In a similar fashion, we aim to build a generator that trains with image patches, and inference images of unbounded arbitrary-large size.
%\hubert{inferences images well beyond its training data size, and generalizes to an unbounded arbitrarily-large size.}
% and at inference time synthesizes images well beyond its training data size.
%
% The generator can thus generalize to an unbounded arbitrarily-high resolution.
% Large-resolution images are at a patch-based manner.
%generated one patch at a time.
%At inference time, the generator synthesizes images well beyond its training data resolution, generalizing to an unbounded arbitrarily-high resolution.
% Motivated by and analogous to the human capability, we aim to train on small easily obtainable patches and at inference time can extend arbitrarily beyond image resolution seen at training time. 
%Based on such reasoning, our method trains on small easily obtainable patches and at inference time can extend arbitrarily beyond image resolution seen at training time. 
%
An example of a synthesized scene containing globally-plausible structure and heterogeneous textures is shown in \figref{teaser}.

We propose InfinityGAN, a method that trains on a finite-pixel dataset, while generating infinite-pixel images at inference time.
% We assume four disentangled factors: \textit{global appearances} shared among all patches for holistic appearance, \textit{coordinates} designating the target patches, and \textit{local structures} and \textit{local textures} modeling local details.
% 
InfinityGAN consists of a neural implicit function, termed \emph{structure synthesizer}, and a padding-free StyleGAN2 generator, dubbed \emph{texture synthesizer}. 
Given a global appearance of an infinite-pixel image, the structure synthesizer samples a sub-region using coordinates and synthesizes an intermediate local structural representations.
% local structures to obtain local structural representations.
The texture synthesizer then seamlessly synthesizes the final image by parts after filling the fine local textures to the local structural representations.
InfinityGAN can infer a compelling global composition of a scene with realistic local details.
% it locally. 
Trained on small patches, InfinityGAN achieves high-quality, seamless and arbitrarily-sized outputs with low computational resources---a single TITAN X to train and test.

We conduct extensive experiments to validate the proposed method. Qualitatively, we present the everlastingly long landscape images. Quantitatively, we evaluate InfinityGAN and related methods using user study and a proposed ScaleInv FID metric. Furthermore, we demonstrate the efficiency and efficacy of the proposed methods with several applications. First, we demonstrate the flexibility and controllability of the proposed method by spatially fusing structures and textures from different distributions within an image. Second, we show that our model is an effective deep image prior for the image outpainting task with the image inversion technique and achieves multi-modal outpainting of arbitrary length from arbitrarily-shaped inputs. Third, with the proposed model we can divide-and-conquer the full image generation into independent patch generation and achieve 7.2$\times$ of inference speed-up with parallel computing, which is critical for high-resolution image synthesis.

%% file: 2-related_work.tex
\input{tex/3-fig_arch}
\vskip \secmargin
\vspace{-1.5mm}
\vspace{\secmargin}
\section{Related Work}
\vspace{\secmargin}
\paragraph{Latent generative models.}

%Regardless of the underlying frameworks, 
Existing generative models are mostly designed to synthesize images of fixed sizes. 
A few methods~\citep{karras2018progressive,karras2020analyzing} have been recently developed to train latent generative models on high-resolution images, up to 1024$\times$1024 pixels.
However, latent generative models generate images from dense latent vectors that require synthesizing all structural contents at once.
Bounded by computational resources and limited by the learning framework and architecture, these approaches synthesize images of certain sizes and are non-trivial to generalize to different output size.
In contrast, patch-based GANs trained on image  patches~\citep{lin2019cocogan,shaham2019singan,shocher2019ingan} are less constrained by the resource bottleneck with the synthesis-by-part approach. 
However, \citep{shaham2019singan,shocher2019ingan} can only model and repeat internal statistics of a single image, and \citep{lin2019cocogan} can barely extrapolate few patches beyond the training size.
% 
% \hubert{MS-PIE~\citep{xu2021positional} proposes to improve SinGAN with sinusoidal coordinate encoding, we still observe the method fail at larger output sizes.}
% The effect of positional information is also visited in MS-PIE~\citep{xu2021positional}, however, it mainly focuses on the coordinate interpolation cases instead of extending the generated structure.}
% 
ALIS~\citep{skorokhodov2021aligning} is a concurrent work that also explores synthesizing infinite-pixel images. It recursively inbetweens latent variable pairs in the horizontal direction. We further discuss the method in Appendix A.
% 
% Finally, autoregressive models~\citep{oord2016pixel,razavi2019vqvae2,esser2021taming} can theoretically synthesize at arbitrary image sizes. The potential of extending these models to infinite-pixel image synthesis task has not yet been explored.
Finally, autoregressive models~\citep{oord2016pixel,razavi2019vqvae2,esser2021taming} can theoretically synthesize at arbitrary image sizes. Despite \citep{razavi2019vqvae2} and \citep{esser2021taming} showing unconditional images synthesis at 1024$\times$1024 resolution, their application in infinite-pixel image synthesis has not yet been well-explored.

% may also achieve infinitely extended synthesis results. However, its pixel-by-pixel synthesis paradigm results in subpar visual quality.
% Some recent works~\citep{razavi2019vqvae2,esser2021taming} combine PixelCNN with CNN-decoders to improve visual quality. However, it also introduces zero-paddings with incorrect positional information,

% with a similar approach in sequentially synthesizing the intermediate quantized features adopt a CNN-decoder instead of pixel-by-pixel decoding. But the use of CNN-decoder introduces zero-paddings with incorrect positional information, thus suffering from the similar problems of SinGAN and StyleGAN baselines we discussed in Section~\ref{exp:qualitative}.

\vspace{-1mm}
\vspace{\paramargin}
\paragraph{Conditional generative models.}
Numerous tasks such as image super-resolution, semantic image synthesis, and image extrapolation often showcase results over 1024$\times$1024 pixels.
These tasks are less related to our setting, as most structural information is already provided in the conditional inputs.
We illustrate and compare the characteristics of these tasks against ours  in Appendix B.

\vspace{-1mm}
\vspace{\paramargin}
\paragraph{Image outpainting.}
Image outpainting~\citep{abdal2020image2styleganpp,liu2020infinite,sabini2018outpainting,yang2019very} is related to image inpainting~\citep{liu2018image,yu2018free} and shares similar issues that the generator tends to copy-and-paraphrase the conditional input or create mottled textural samples, leading to repetitive results especially when the outpainted region is large. 
%
%Thus, all these methods sequentially fills-in outpainting area when the region is too large, but still yields repetitive results.
%
InOut~\citep{inout} proposes to outpaint image with GANs inversion and yield results with higher diversity.
We show that with InfinityGAN as the deep image prior along with InOut \citep{inout}, we obtain the state-of-the-art outpainting results and avoids the need of sequential outpainting.
Then, we demonstrate applications in arbitrary-distant image inbetweening, which is at the intersection of image inpainting~\citep{liu2018image,Nazeri_2019_ICCV,yu2018free} and outpainting research.

\vspace{-1mm}
\vspace{\paramargin}
\paragraph{Neural implicit representation.} 
Neural implicit functions~\citep{part2019deepsdf,mescheder2019occupancy,mildenhall2020nerf} have been applied to model the structural information of 3D and continuous representations.
Adopting neural implicit modeling, our query-by-coordinate synthesizer is able to model structural information effectively.
Some recent works~\citep{devries2021unconstrained,niemeyer2021giraffe,chan2021pi} also attempt to integrate neural implicit function into generative models, but aiming at 3D-structure modeling instead of extending the synthesis field-of-view.
 % A reviewer complained about this, though I don't really get the point. I guess he thought using implicit function is the main novelty.
% 
% \hubert{
% In particular, LIIF~\citep{chen2020liif} achieves infinite-zoom-in effect by interpolating the coordinate condition, which is related but opposite to our extension-oriented task.
% }

%% file: tex/3-fig_arch.tex
\begin{figure}[t!]
    \centering
    \includegraphics[width=0.98\linewidth]{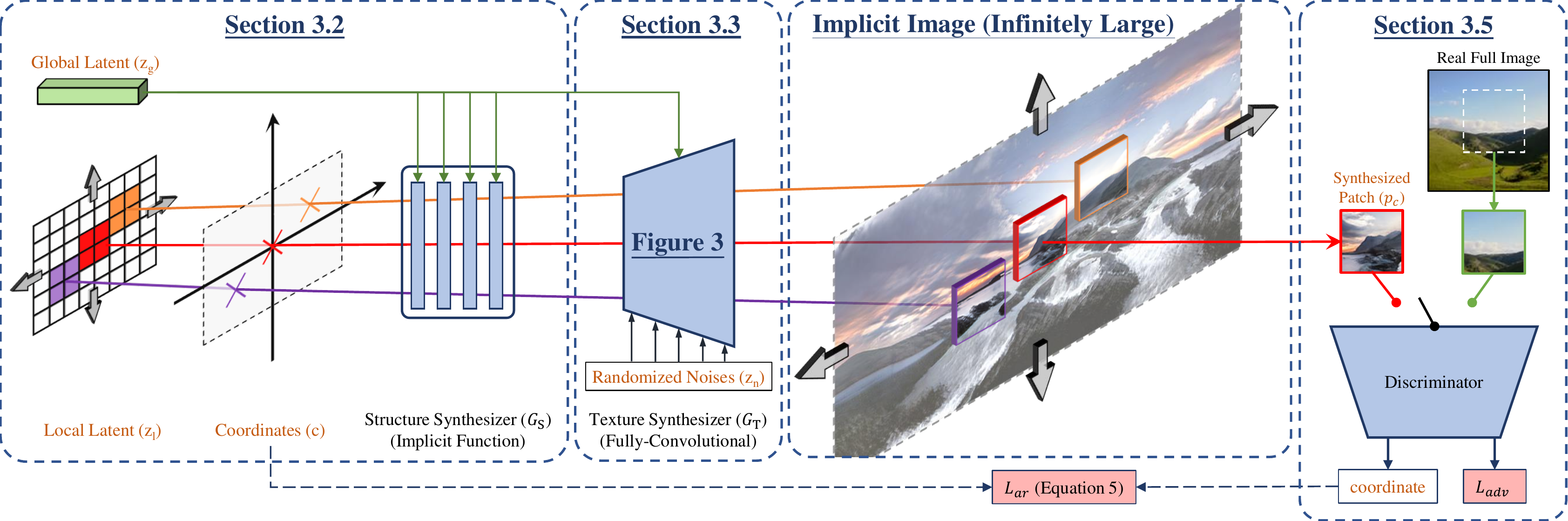}
    \vspace{-3mm}
    \caption{
    \textbf{Overview.}
    The generator of InfinityGAN consists of two modules, a structure synthesizer based on a neural implicit function, and a fully-convolutional texture synthesizer with all positional information removed (see Figure~\ref{fig:fully_conv_generator}).
    The two networks take four sets of inputs, a global latent variable that defines the holistic appearance of the image, a local latent variable that represents the local and structural variation, a continuous coordinate for learning the neural implicit structure synthesizer, and a set of randomized noises to model fine-grained texture.
    %
    %With a standard discriminator, all modules are end-to-end trained with adversarial learning.
    %
    InfinityGAN synthesizes images of arbitrary size by learning spatially extensible representations.
    %\hy{is it possible to make the text a bit larger, I can barely see the annotation (like zn). For example, tilt the implicit image a bit more to spare more space as it provide limited info. }
    %
    }
    \vspace \figmargin
    \vspace{-1.2mm}
    \label{fig:overview}
\end{figure}

%% file: 3-method.tex
\vspace{\secmargin}
%\vspace{-0mm}
\section{Proposed Method}
\vspace{\secmargin}

\subsection{Overview}
\vspace{\subsecmargin}

An arbitrarily large image can be described globally and locally. 
Globally, images should be coherent and hence global characteristics should be expressible by a compact \emph{holistic appearance} (\eg a medieval landscape, ocean view panorama). Therefore, we adopt a fixed holistic appearance for each infinite-pixel image to represent the high-level composition and content of the scene. 
Locally, a close-up view of an image is defined by its local structure and texture.
The structure represents objects, shapes and their arrangement within a local region. 
Once the structure is defined, there exist multiple feasible appearances or textures to render realistic scenes. 
% 
% Hence, texture synthesis can be seen as a second step, conditioned on the structure, material properties and lighting.
%
At the same time, structure and texture should conform to the global \emph{holistic appearance} to maintain the visual consistency among the neighboring patches. 
Given these assumptions, we can generate an infinite-pixel image by first sampling a global holistic appearance, then spatially extending local structures and textures following the holistic appearance.  
%an infinite-pixel image can be generated accordingly.
%Once the global holistic appearance is sampled, local structure and texture can be spatially extended infinitely as long as they follow holistic appearance.

Accordingly, the InfinityGAN generator $G$ consists of a \emph{structure synthesizer} $\structuresyn$ and a \emph{texture synthesizer} $\texturesyn$. 
$\structuresyn$ is an implicit function that samples a sub-region with coordinates and creates local structural features.
$\texturesyn$ is a fully convolutional StyleGAN2~\citep{karras2020analyzing} modeling textural properties for local patches and rendering final image.
Both modules follow a consistent holistic appearance throughout the process.
% The former is an implicit function that models the global holistic appearance and local structure of the image. The latter is a fully convolutional StyleGAN2~\citep{karras2020analyzing} modeling the textural properties for local patches while still attended to the global holistic appearance.
% 
\figref{overview} presents the overview of our framework.

\vspace{-0.5mm}
\subsection{Structure Synthesizer ($\structuresyn$)}
\label{method:ss}
\vspace{\subsecmargin}

Structure synthesizer is a neural implicit function driven by three sets of latent variables:
A global latent vector $\globalz$ representing the holistic appearance of the infinite-pixel image (also called \emph{implicit image} since the whole image is never explicitly sampled), a local latent tensor $\localz$ expressing the local structural variation of the image content, and a coordinate grid $\coordgrid$ specifying the location of the patches to sample from the implicit image.
The synthesis process is formulated as:
\vspace{\eqmargin}
\begin{equation}
    \structurez = \structuresyn(\globalz, \localz, \coordgrid),
\label{eq:structuresyn}
\vspace{\eqmargin}
\end{equation}
where $\structurez$ denotes the structural latent variable that is later used as an input to the texture synthesizer.

% Given a global latent vector $\globalz$ representing the holistic appearance of the infinite-pixel image (also called implicit image) and a local latent code $\localz$ expressing the local structural variation of patches, structure synthesizer is a neural implicit function that samples a sub-region from the 

% Structure synthesizer is a neural implicit function that aims to sample an implicit representation conditioned on the global $\globalz$ and local $\localz$ latent variables, and generate the structure at the queried location $\coordgrid$. 
% 
% The latent variable $\globalz$ serves as a global appearance descriptor.
%
We sample $\globalz  \in \mathbb{R}^{D_{\globalz}}$ from a unit Gaussian distribution once and inject $\globalz$ into every layer and pixel in $\structuresyn$ via feature modulation~\citep{huang2017adain,karras2020analyzing}.
%
% Local variations of image structure are represented by $\localz$.
%
As local variations are independent across the spatial dimension, we independently sample them from a unit Gaussian prior for each spatial position of $\localz \in \mathbb{R}^{H\times W \times {D_{\localz}}}$, where $H$ and $W$ can be arbitrarily extended.
%forming a three-dimensional tensor that can be arbitrarily extended in the spatial dimensions.
%\hy{so z_l is a infinite 2D grid, }
% \hubert{And the patches as well. I understand your point, but we will need to define a new symbol for both z_l and c for patches. I'm not sure it makes things messy.}
% 
% ~\sergey{why volumetric, isn't it a feature tensor?}
%
% We provide $\localz$ as the input to $\structuresyn$. \hy{A bit ambiguous here. Technically, the queried regions cropped from zl is the input to Gs, given Fig 2. So, does zl mean the uncropped arbitrary-sized latent or the cropped one?} \hubert{that will need to introduce some subscripts to z_l, probably let's omit that for simplicity?}
%
% %  Not important
% Such conditioning of neural implicit function is a less common, but similar formulations can be found in \citep{chen2020liif,saito2019pifu}.%, \hubert{where an implicit function takes encoder predicted local features as a part of its inputs.}

% Finally, with the sampled implicit representation conditioned on $\globalz$ and arbitrarily large $\localz$, the coordinate $\coordgrid$ serves as a query to obtain the region to be retrieved from the implicit image. \sergey{what is implicit image here? we haven't defined it and it's not used afterwards.} \hubert{Moved to the beginning}
%
We then use coordinate grid $\coordgrid$ to specify the location of the target patches to be sampled.
To be able to condition $\structuresyn$ with coordinates infinitely far from the origin, we introduce a prior by exploiting the nature of landscape images: (a) self-similarity for the horizontal direction, and (b) rapid saturation (\eg land, sky or ocean) for the vertical direction.
To implement this, we use the positional encoding for the horizontal axis similar to \citep{vaswani2017attention,tancik2020fourier,sitzmann2020implicit}. We use both sine and cosine functions to encode each coordinate for numerical stability. For the vertical axis, to represent saturation, we apply the tanh function.
Formally, given horizontal and vertical indexes $(i_x, i_y)$ of $\localz$ tensor, we encode them as 
% \vspace{\eqmargin}
% \begin{equation}
$
    \coordgrid = (\tanh(i_y), \cos(i_x / T), \sin(i_x / T)) \, ,
$
% \vspace{\eqmargin}
% \end{equation}
where $T$ is the period of the sine function and $\coordgrid$ controls the location of the patch to generate.

% While coordinates repeat periodically, the variations of $\localz$ avoid repeating structures.
To prevent the model from ignoring the variation of $\localz$ and generating repetitive content by following the periodically repeating coordinates, we adopt a mode-seeking diversity loss~\citep{mao2019mode,DRIT_plus} between a pair of local latent variables $\localzfir$ and $\localzsec$ while sharing the same $\globalz$ and $\coordgrid$:
\vspace{\eqmargin}
\begin{equation}
    % [Hubert] Omit the angular distance here, I don't think that makes a fundamental difference, but reviewers may request an ablation...
    \loss{div} =  \lVert \localzfir - \localzsec \rVert_1  \,
    \,\,  /  \,\,
    \lVert \structuresyn(\globalz, \localzfir, \coordgrid) - \structuresyn(\globalz, \localzsec, \coordgrid) \rVert_1 \, .
    \label{eq:divloss}
    \vspace{\eqmargin}
\end{equation}
%
% Notice that $\globalz$ and $\coordgrid$ are shared between $\localzfir$ and $\localzsec$.

% \sergey{I struggle to understand from here till the end of the section. 
% First of all why feature unfolding? \hubert{It is optional, but improves z_s smoothness.}
% Then, is $f$ a feature between the layers? \hubert{yes, its definition was missing.}
% how is $u$ used afterwards (is it used in the rest of the paper?)} \hubert{No, we define it only for indexing in the equation}
Conventional neural implicit functions produce outputs for each input query independently, which is a pixel in $\localz$ for InfinityGAN.
Such a design causes training instabilities and slows convergence, as we show in Figure 37.
% ~\sergey{provide a citation to support, or say 'according to our experiments'}
We therefore adopt the feature unfolding technique~\citep{chen2020liif} to enable $\structuresyn$ to account for the information in a broader neighboring region of $\localz$ and $\coordgrid$, introducing a larger receptive field.
For each layer in $\structuresyn$, before feeding forward to the next layer, we apply a $k \times k$ feature unfolding transformation at each location $(i,j)$ of the origin input $f$ to obtain the unfolded input $f'$:
%at each location $(i,j)$, a $k \times k$ feature unfolding transforms the origin input $f$ into unfolded input $f'$ with:
\vspace{\eqmargin}
\begin{equation}
    f'_{(i,j)} = \text{Concat}( \{ f{(i+n, j+m)} \}_{n,m \in \{-k/2, k/2\} } ) \, ,
    \vspace{\eqmargin}
\end{equation}
where Concat($\cdot$) concatenates the unfolded vectors in the channel dimension.
%
%
% increase the receptive field of $\structuresyn$ \hubert{by stacking the 7$\times$7 neighboring feature within each layer} to consider the information from a broader area of $\localz$ and $\coordgrid$.
%
% The use of feature unfolding turns $\coordgrid$ into a grid of coordinates rather than a simple triplet. \hy{Does it mean, it equivalently turn c into grids or it actually do so in the first layer?}
%
% This approach also helps $\structuresyn$ to process and reflect the interaction among all $\localz$ in its neighborhood.
%
% \hubert{Notice that $\localz$ is magnitudes smaller and more compact than the output image in the spatial dimension, its unfolded \textit{neighbor} can equivalently span through a thousand of pixels in the image space easily without a deep $\structuresyn$.}
% Notice that $\localz$ is a highly compressed representation before decoded into the image space, thus its \textit{neighbor} can easily span through a thousand of pixels in the image space with a shallow $\structuresyn$.
%
In practice, as the grid-shaped $\localz$ and $\coordgrid$ are sampled with equal spacing between consecutive pixels, the feature unfolding can be efficiently implemented with CoordConv~\citep{liu2018coordconv}.

%HERE
\vspace{-0.5mm}
\subsection{Texture Synthesizer ($\texturesyn$)}
\vspace{\subsecmargin}
\label{method:ts}
%
%MH: this sentence can be further improved to make it more novel (not just replace one component of StyleGAN2). 
%Hubert: I believe we should describe these modifications more like reasonable adaptations. The novelty is not in this part. I don't want to see reviewers request useless ablations in this part.
Texture synthesizer aims to model various realizations of local texture given the local structure $\structurez$ generated by the structure synthesizer. 
% \sergey{Formally define $\texturesyn$ : inputs $\longrightarrow$ outputs}
In addition to the holistic appearance $\globalz$ and the local structural latent $\structurez$, texture synthesizer uses noise vectors $\noisez$ to model the finest-grained textural variations that are difficult to capture by other variables.
The generation process can be written as:
\vspace{\eqmargin}
\begin{equation}
    \patch = \texturesyn({\structurez, \globalz, \noisez}) \, ,
\vspace{\eqmargin}
\end{equation}
where $\patch$ is a generated patch at location $\coordgrid$ (\ie the $\coordgrid$ used in Eq~\ref{eq:structuresyn} for generating $\structurez$).

We implement upon StyleGAN2~\citep{karras2020analyzing}.
First, we replace the fixed constant input with the generated structure $\structurez$. 
% \hy{dimension of zS, is it a single vector or a spatial tensor} \hubert{probably does not effect reading here?}
% 
Similar to StyleGAN2, randomized noises $\noisez$ are added to all layers of $\texturesyn$, representing the local variations of fine-grained textures.
% To model local variations of texture we use pass $\noisez$ that injects randomized noises to model fine-grained random texture. \sergey{how is it injected?}
%
Then, a mapping layer
% , implemented with multi-layer perceptrons, 
projects $\globalz$ to a style vector, and the style is injected into all pixels in each layer via feature modulation.
Finally, we remove all zero-paddings from the generator, as shown in \figref{fully_conv_generator}(b). 

\input{tex/3-fig_fully_conv_generator}

Both zero-padding and $\structuresyn$ can provide positional information to the generator, and we later show that positional information is important for generator learning in Section~\ref{exp:ablation}.
% we observe StyleGAN2 generator relying on zero-padding to understand the location of the pixel-to-synthesize in the output image.
However, it is necessary to remove all zero-paddings from $\texturesyn$ for three major reasons.
First, zero-padding has a consistent pattern during training, due to the fixed training image size.
Such a behavior misleads the generator to memorize the padding pattern, and becomes vulnerable to unseen padding patterns while attempting to synthesize at a different image size.
The third column of Figure~\ref{fig:exp-qualitative} shows when we extend the input latent variable of the StyleGAN2 generator multiple times, the center part of the features does not receive expected coordinate information from the paddings, resulting in extensively repetitive textures in the center area of the output image.
Second, zero-paddings can only provide positional information within a limited distance from the image border.
However, while generating infinite-pixel images, the image border is considered infinitely far from the generated patch.
Finally, as shown in Figure~\ref{fig:fully_conv_generator}, the existence of paddings hampers $\texturesyn$ from generating separate patches that can be composed together.
Therefore, we remove all paddings from $\texturesyn$, facilitating the synthesis-by-parts of arbitrary-sized images.
We refer to the proposed $\texturesyn$ as a padding-free generator (PFG).

%HERE
\vspace{-1mm}
\subsection{Spatially Independent Generation}
\vspace{\subsecmargin}
\label{method:spatial-independent}
InfinityGAN enables spatially independent generation thanks to two characteristics of the proposed modules.
First, $\structuresyn$, as a neural implicit function, naturally supports independent inference at each spatial location.
Second, $\texturesyn$, as a fully convolutional generator with all paddings removed, can synthesize consistent pixel values at the same spatial location in the implicit image, regardless of different querying coordinates, as shown in Figure~\ref{fig:fully_conv_generator}(b).
%As shown in Figure~\ref{fig:fully_conv_generator}(b), the removal of all paddings makes $\texturesyn$ a fully convolutional generator that synthesizes consistent pixel values at the same spatial location in the implicit image, regardless of the selection of $\structurez$ slices.
% 
% Meanwhile, the size of the $\structurez$ generated with $\structuresyn$ is aligned to $\localz$ and $\coordgrid$ as implicit functions do not alter the spatial dimension with up- or down-sampling.
% %
% With these properties, we can form a one-to-one mapping from each pixel in the generated images, to a set of input latent variables that contribute to that pixel.
% %
% That is, with proper indexing in $\localz$, $G$ will synthesize a consistent value at each pixel location in the image, regardless of the order of the patches are generated.
% %
% This characteristic enables the model to independently generate patches that can be combined seamlessly, and maintains a \textit{constant} memory usage while synthesizing images of any size.
% 
%Meanwhile, $\structuresyn$ as a neural implicit function can naturally inference at each spatial location independently.
% 
With these properties, we can independently query and synthesize a patch from the implicit image, seamlessly combine multiple patches into an arbitrarily large image, and maintain \textit{constant} memory usage while synthesizing images of any size.

In practice, having a single center pixel in a $\structurez$ slice that aligns to the center pixel of the corresponding output image patch can facilitate $\localz$ and $\coordgrid$ indexing.
We achieve the goal by shrinking the StyleGAN2 blur kernel size from $4$ to $3$, causing the model to generate odd-sized features in all layers, due to the convolutional transpose layers.

\vspace{-1mm}
\subsection{Model Training}
\vspace{\subsecmargin}
\label{section:method-training}
% Different from COCO-GAN, we do not need a D for larger FoV
% Describe the roles of patches and full images
% Describe hyperparameters
%Although the output resolution of $G$ is a bit different from StyleGAN2, 
%MH: bad sentence
%We use the same $D$ architecture of StyleGAN2.
The discriminator $D$ of InfinityGAN is similar to the one in the StyleGAN2 method.
The detailed architectures of $G$ and $D$ are presented in Appendix D.
The two networks are trained with the non-saturating logistic loss $\loss{adv}$~\citep{goodfellow2014generative}, $R_1$ regularization $\loss{R_1}$~\citep{mescheder2018r1reg} and path length regularization $\loss{path}$~\citep{karras2020analyzing}.
Furthermore, to encourage the generator to follow the conditional distribution in the vertical direction, we train $G$ and $D$ with an auxiliary task~\citep{odena2017acgan} predicting the vertical position of the patch:
\vspace{\eqmargin}
\begin{equation}
    \loss{ar} = \lVert \hat{\coordgrid}_y - \bar{\coordgrid}_y \rVert_1 \, ,
    \vspace{\eqmargin} 
    \label{eq:arloss}
\end{equation}
where $\hat{\coordgrid}_y$ is the vertical coordinate predicted by $D$, and $\bar{\coordgrid}_y$ is either (for generated images) $\coordgrid_y = \tanh(i_y)$ or (for real images) the vertical position of the patch in the full image.
We formulate $\loss{ar}$ as a regression task. % , and correspondingly, we use the center of $\coordgrid$ for $\bar{\coordgrid}$ in Equation~\eqref{eq:arloss}.
The overall loss function for the InfinityGAN is:
\vspace{\eqmargin}
\begin{equation}
\setlength{\jot}{-3pt} % Reduces the distance between aligned equations
\begin{aligned}
    % \min_D \, \, \max_G \, \,  & \left[ L_{adv} + \lambda_{R_1} L_{R_1} + \lambda_{path} L_{path} \right] \, + \\
    % \min_D \, \, \min_G \, \, & \lambda_{ar} L_{ar} \, + \, \min_G \, \,  \lambda_{div}L_{div}
    \min_D \, \, \, &   \loss{adv} + \lambdaw{ar} \loss{ar} + \lambdaw{R_1} \loss{R_1} \, \, , \\
    \min_G \, \, \, & - \loss{adv} + \lambdaw{ar} \loss{ar} + \lambdaw{div}\loss{div} + \lambdaw{path}\loss{path} \,\, ,
\end{aligned}
\vspace{\eqmargin}
\end{equation}
where $\lambda$'s are the weights.

%
%------------------------------------------------------------------------

%% file: tex/3-fig_fully_conv_generator.tex
\begin{figure*}[t!]
    \centering
    \includegraphics[width=\linewidth]{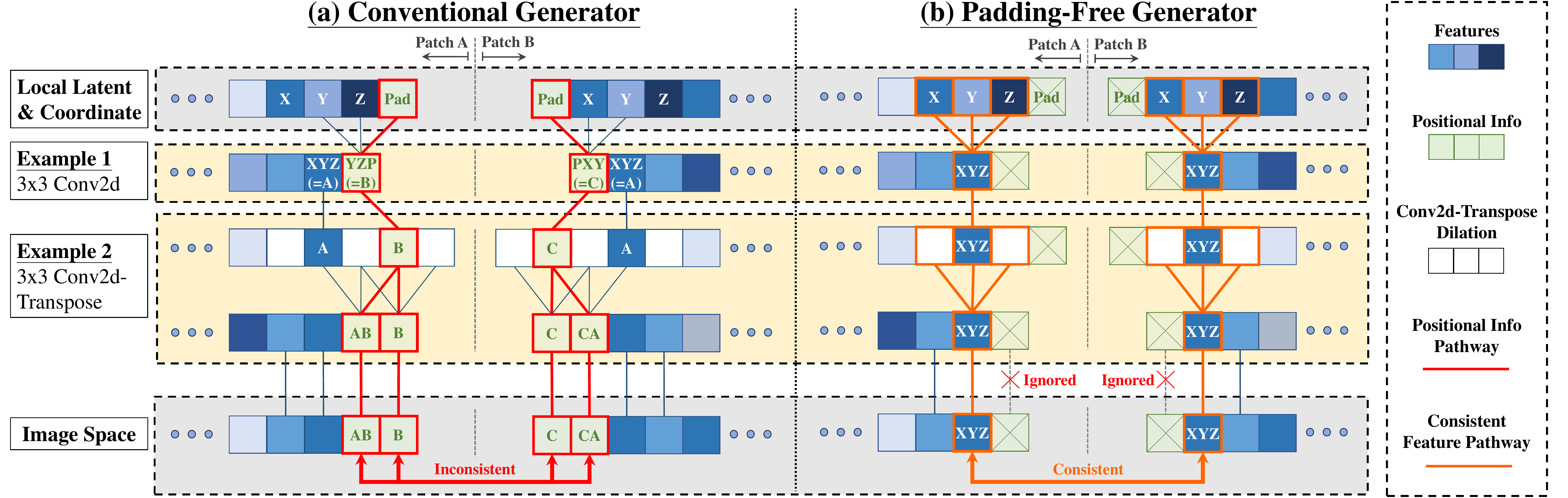}
    \vspace{-6mm}
    \caption{
    \textbf{Padding-free generator.}
    (Left) Conventional generators synthesize inconsistent pixels due to the zero-paddings. Note that the inconsistency region grows exponentially as the network deepened. (Right) In contrast, our padding-free generator can synthesize consistent pixel value regardless of the position in the model receptive field. Such a property facilitates spatially-independently generating patches and forming into a seamless image with consistent feature values.
    % To make the texture synthesizer learn only texture representations rather than structural information, we remove all paddings from the generator to avoid position encoding.
    % %
    % With such a fully-convolutional generator design, we can spatial-independently generate patches and forming into a seamless image.
    %
    }
    \vspace \figmargin
    \label{fig:fully_conv_generator}
\end{figure*}

%% file: 4-experiment.tex
\input{tex/4-tab_quant}
\input{tex/4-fig_quali}

% \vspace \secmargin
\section{Experimental Results}
\vspace \secmargin

%\vspace{\paramargin}
\noindent\textbf{Datasets.} 
%
%MH: not testify... "testify" is used in court
%We mainly testify our method on Flickr-Landscape, a dataset randomly crawled from the landscape group on Flickr, consists of $50{,}000$ high-quality landscape images. 
We evaluate the ability of synthesizing at extended image sizes on the Flickr-Landscape dataset consists
of $450{,}000$ high-quality landscape images, which are 
crawled from the Landscape group on Flickr. 
For the image outpainting experiments, we evaluate with other baseline methods on scenery-related subsets from the Place365~\citep{zhou2017places} ($62{,}500$ images) and Flickr-Scenery~\citep{inout} ($54{,}710$ images) datasets.
Note that the Flickr-Scenery here is different from our Flickr-Landscape.
For image outpainting task, we split the data into $80\%$, $10\%$, $10\%$ for training, validation, and test.
All quantitative and qualitative evaluations are conducted on test set.
% We present the quantitative and qualitative results on the test set only. 
%
%\hubert{I will make this line in the first page (somehow it was removed...)}
%The source code, trained models, and annotated datasets will be made available to the public. 

%\vspace{\paramargin}
\noindent\textbf{Hyperparameters.} 
%
%For InfinityGAN training, we use $\lambda_{ar}=1$, $\lambda_{div}=1$, $\lambda_{R_1}=10$, and $\lambda_{path}=2$ for all datasets. 
We use $\lambdaw{ar}=1$, $\lambdaw{div}=1$, $\lambdaw{R_1}=10$, and $\lambdaw{path}=2$ for all datasets.
All models are trained with 101$\times$101 patches cropped from 197$\times$197 real images.
Since our InfinityGAN synthesizes odd-sized images, we choose 101 that maintains a sufficient resolution that humans can still recognize its content. On the other hand, 197 is the next output resolution if stacking another upsampling layer to InfinityGAN, which also provides 101$\times$101 patches a sufficient field-of-view.
We adopt the Adam~\citep{adam} optimizer with $\beta_1=0$, $\beta_2=0.99$ and a batch size 16 for 800,000 iterations.
More details are presented in Appendix C.

%\vspace{\paramargin}
\noindent\textbf{Metrics.}
We first evaluate Fr\'echet Inception Distance (\textbf{FID})~\citep{fid} at $G$ training resolution.
%to show the performance and expressiveness of $G$ at its training distribution.
%
Then, without access to real images at larger sizes, we assume that the real landscape with a larger FoV will share a certain level of self-similarity with its smaller FoV parts.
We accordingly propose a \textbf{ScaleInv FID}, which resizes larger images to the training data size with bilinear interpolation, then computes FID.
We denote N$\times$ ScaleInv FID when the metric is evaluated with images N$\times$ larger than the training samples.
% We denote the ScaleInv FID evaluated with N$\times$ resolution (before resizing) as N$\times$ ScaleInv FID afterward.

%\vspace{\paramargin}
\noindent\textbf{Evaluated Method.}
We perform the evaluation on Flickr-Landscape with the following algorithms:
\begin{compactitem}[$\bullet$]
    \item[$-$]\textbf{SinGAN.}
We train an individual SinGAN model for each image.
% , resulting in $50{,}000$ SinGAN models.
%
The images at larger sizes are generated by setting spatially enlarged input latent variables.
Note that we do not compare with the super-resolution setting from SinGAN since we focus on extending the learned structure rather than super-resolve the high-frequency details.
% it heavily borrows the structural information from the training image with memorization.
\item[$-$]\textbf{COCO-GAN.}
Follow the ``Beyond-Boundary Generation'' protocol of COCO-GAN, we transfer a trained COCO-GAN model to extended coordinates with a post-training procedure.
% we transfer the learned representations from a trained standard COCO-GAN to extended coordinates with a post-training procedure.
%
\item[$-$]\textbf{StyleGAN2 (+NCI).}
We replace the constant input of the original StyleGAN2 with a $\localz$ of the same shape, we call such a replacement as ``non-constant input (NCI)''.
This modification enables StyleGAN2 to generate images at different output sizes with different $\localz$ sizes.
\end{compactitem}

\vspace{-3mm}
\subsection{Generation at Extended Size.}
\label{exp:quantitative}
\vspace{\subsecmargin}

%\vspace{\paramargin}
%\noindent\textbf{Extra unfair protocols for fairness.}
\noindent\textbf{Additional (unfair) protocols for fairness.}
We adopt two additional pre- and post-processing to ensure that InfinityGAN does not take advantage of its different training resolution.
To ensure InfinityGAN is trained with the same amount of information as other methods, images are first bilinear interpolated into 128$\times$128 before resized into 197$\times$197.
Next, for all testing sizes in Table~\ref{fig:exp-qualitative}, InfinityGAN generates at 1.54$\times$ ($=$197$/$128) larger size to ensure final output images share the same FoV with others.
In fact, these corrections make the setting disadvantageous for InfinityGAN, as it is trained with patches of 50\% FoV, generates 54\% larger images for all settings, and requires to composite multiple patches for its 1$\times$ ScaleInv FID.
%
%Despite the constraints, as shown in Table~\ref{tab:exp-quantitative}, InfinityGAN starts outperform all baselines starting from 4$\times$ ScaleInv FID, and over 90\% more favorable than all baselines from the user study.

%\vspace{\paramargin}
\noindent\textbf{Quantitative analysis.}
For all the FID metrics in Table~\ref{tab:exp-quantitative}, unfortunately, the numbers are not directly comparable, since InfinityGAN is trained with patches with smaller FoV and at a different resolution.
Nevertheless, the trend in ScaleInv FID is informative.
It reflects the fact that the global structures generated from the baselines drift far away from the real landscape as the testing FoV enlarges.
Meanwhile, InfinityGAN maintains a more steady slope, and surpasses the strongest baseline after 4$\times$ ScaleInv FID.
Showing that InfinityGAN indeed performs favorably better than all baselines as the testing size increases.

\input{tex/4-fig_lsun_and_diversity}

%\vspace{\paramargin}
\noindent\textbf{Qualitative results.}
\label{exp:qualitative}
In Figure~\ref{fig:exp-qualitative}, we show that all baselines fall short of creating reasonable global structures with spatially expanded input latent variables.
%
% SinGAN mainly generates a large amount of textural content in the image center while synthesizing at extended sizes.
% %
COCO-GAN is unable to transfer to new coordinates when the extrapolated coordinates are too far away from the training distribution.
Both SinGAN and StyleGAN2 implicitly establish image features based on position encoded by zero padding, assuming the training and testing position encoding should be the same.
However, when synthesizing at extended image sizes, the inevitable change in the spatial size of the input and the features leads to drastically different position encoding in all model layers.
Despite the models can still synthesize reasonable contents near the image border, where the position encoding is still partially correct, they fail to synthesize structurally sound content in the image center.
% SinGAN memorizes the structural information with position encoding from zero padding, and tends to place all memorized objects near the border of the images.
%
%
% As for StyleGAN2, the extended input $\localz$ for generation at the extended sizes introduces different padding patterns from training, causing StyleGAN2 fails to generate reasonable global structures but generates texture-like patterns.
%
Such a result causes ScaleInv FID to rapidly surge as the extended generation size increases to 1024$\times$1024.
Note that at the 16$\times$ setting, StyleGAN2 runs out of memory with a batch size of 1 and does not generate any result.
In comparison, InfinityGAN achieves reasonable global structures with fine details.
Note that the 1024$\times$1024 image from InfinityGAN is created by compositing 121 independently synthesized patches.
With the ability of generating consistent pixel values (Section~\ref{method:spatial-independent}), the composition is guaranteed to be seamless.
We provide more comparisons in Appendix E, a larger set of generated samples in Appendix F, results from models trained at a higher resolution in Appendix G, and a very-long synthesis result in Appendix J.

In Figure~\ref{fig:lsun}, we further conduct experiments on LSUN bridge and tower datasets, demonstrating InfinityGAN is applicable on other datasets.
However, since the two datasets are object centric with a low view-angle variation in the vertical direction, InfinityGAN 
frequently fills the top and bottom area with blank padding textures.
% learns to fill-in the top and bottom area with blurry contents.

% \input{tex/4-fig_diversity}
In Figure~\ref{fig:diversity}, we switch different $\localz$ and $\texturesyn$ styles (\ie $\globalz$ projected with the mapping layer) while sharing the same $\coordgrid$. More samples can be found in Appendix I.
The results show that the structure and texture are disentangled and modeled separately by $\structuresyn$ and $\texturesyn$.
The figure also shows that $\structuresyn$ can generate a diverse set of structures realized by different $\localz$.

%\vspace{\paramargin}
\noindent\textbf{User study.}
% 
% We ask users to select images with more favorable global structures from two sets of $16$ images generated from two competing methods.
% %
% The evaluation is conducted by pairing our InfinityGAN against each of the other baselines.
% %
% We ask each user to compare 30 pairs and collect the results from 50 participants.
% %
% As presented in Table~\ref{tab:exp-quantitative}, the user study shows an over 90\% of preference favorable to InfinityGAN against all other baselines.
% 
We use two-alternative forced choice (2AFC) between InfinityGAN and other baselines on the Flickr-Landscape dataset.
A total of 50 participants with basic knowledge in computer vision engage the study, and we conduct 30 queries for each participant.
For each query, we show two separate grids of 16 random samples from each of the comparing methods and ask the participant to select ``the one you think is more realistic and overall structurally plausible.''
As presented in Table~\ref{tab:exp-quantitative}, the user study shows an over 90\% of preference favorable to InfinityGAN against all baselines.
%
%Despite the ranking order of the methods is approximately aligned with ScaleInv FID, the close quantitative values from ScaleInv FID fails to correctly reflect the significance in performance difference.
\input{tex/4-fig_spatial_fusion}

% \vspace{\subsecmargin}
\vspace{\subsecmargin}
\subsection{Ablation Study: The Positional Information In Generator}
\label{exp:ablation}
\vspace{\subsecmargin}
As discussed in Section~\ref{method:ts}, we hypothesize that StyleGAN2 highly relies on the positional information from the zero-paddings.
% memorizes the structure of generated contents with position encoding.
%
% \sergey{NCI, PFG should be introduce here as well}
% \hubert{They have been named in Section 4 and Section 3.3, respectively}
In Table~\ref{tab:exp-quantitative} and Figure~\ref{fig:exp-qualitative}, we perform an ablation by removing all paddings from StyleGAN2+NCI, yielding StyleGAN2+NCI+PFG that has no positional information in the generator.
The results show that StyleGAN2+NCI+PFG fails to generate reasonable image structures, and significantly degrades in all FID settings.
Then, with the proposed $\structuresyn$, the positional information is properly provided from $\structurez$, and resumes the generator performance back to a reasonable state.

\input{tex/4-tab_outpaint_quant_and_speed_bench}
\input{tex/4-fig_outpaint_quali}
\input{tex/4-fig_inbetweening}

\vspace{\subsecmargin}
\subsection{Applications}
\vspace{\subsecmargin}

%\vspace{\paramargin}
\noindent\textbf{Spatial style fusion.}
\label{exp:spatial-fusion}
Given a single global latent variable $\globalz$, the corresponding infinite-pixel image is tied to a single modal of global structures and styles.
To achieve greater image diversity and allow the user to interactively generate images, 
we propose a spatial fusion mechanism that can spatially combine two global latent variables with a smooth transition between them.
%Synthesizing an infinite-pixel image with a single global latent variable will result in a less interesting global view and interactiveness, since all the local structures and textures are sampled from a minor set of distribution confined by $\globalz$.
%
%We accordingly propose a spatial fusion mechanism that effectively fuses two global latent variables in the spatial dimension with smooth transition.
%
First, we manually define multiple style centers in the pixel space and then construct an initial fusion map by assigning pixels to the nearest style center.
The fusion map consists of one-hot vectors for each pixel, forming a style assignment map. 
According to the style assignment map, we then propagate the styles in all intermediate layers. Please refer to Appendix L for implementation details.
% We then propagate the style assignments to all intermediate layers.
% \sergey{I don't understand this $\longrightarrow$}Then, we conduct a reverse procedure to yield fusion maps for every intermediate features.
%
%At the beginning, we first specify the style centers in the pixel space.
%
%An initial fusion map can be constructed by pixel-wise calculating the closest style center, and store the information as one-hot vectors for each pixel, forming a volumetric tensor.
%
%Here we may optionally add a Gaussian blur with large kernel size, which increases the fusion visual smoothness.
%
%Then, we reverse engineer the change in spatial shape of intermediate features caused by convolution operations, and yield fusion maps for every intermediate features.
%
%For instance, we use a bilinear interpolation in reverse the effect of a upsampling convolutional transpose.
%
%
Finally, with the fusion maps annotated for every layer, we can apply the appropriate $\globalz$ from each style center to each pixel using feature modulation.% as feature modulation is a pixel-wise operator.
Note that the whole procedure has a similar inference speed as the normal synthesis.
%
%Despite the initial fusion map creation consumes extra time for a large amount of memory allocations, it is a one-time cost that can be shared over samples and batches.
%
Figure~\ref{fig:spatial_fusion} shows synthesized fusion samples.

%\vspace{\paramargin}
\noindent\textbf{Outpainting via GAN Inversion.} 
\label{exp:outpainting}
We leverage the pipeline proposed in \inout~\citep{inout} to perform image outpainting with latent variable inversion.
All loss functions follow the ones proposed in \inout.
We first obtain inverted latent variables that generates an image similar to the given real image via GAN inversion techniques, then outpaint the image by expanding $\localz$ and $\noisez$ with their unit Gaussian prior.
% We first obtain inverted latent variables $\globalz'$, $\localz'$ and $\texturez'$ that generates an image similar to the given real image, then outpaint the image by expanding $\localz'$ to $\hat{\localz}'$ with its unit Gaussian prior.
%
See Appendix K for implementation details.
%
% We first obtained inverted latent variables $\localz'$ and $\globalz'$, then expand them to larger $\hat{\localz}$ and $\hat{\globalz}$ for the target resolution.

In Table~\ref{tab:exp-outpainting-quant}, our model performs favorably against all baselines in image outpainting (Boundless~\citep{teterwak2019boundless}, NS-outpaint~\citep{yang2019very}, 
%Image2StyleGAN++~\citep{Islam2020How}, 
and \inout~\citep{inout}).
% and inpainting (DeepFillv2~\citep{yu2018free}).
%
As shown in \figref{outpaint-quali}, while dealing with a large outpainting area (\eg panorama), all previous outpainting methods adopt a sequential process that generates a fixed region at each step.
This introduces obvious concatenation seams, and tends to produce repetitive contents and black regions after the multiple steps.
In contrast, with InfinityGAN as the image prior in the pipeline of~\citep{inout}, we can directly outpaint arbitrary-size target region from inputs of arbitrary shape.
Moreover, in Figure~\ref{fig:outpaint-multimodal}, we show that our outpainting pipeline natively supports multi-modal outpainting by sampling different local latent codes in the outpainting area.
% or spatial fusion with a different $\hat{\globalz}$.

%\vspace{\paramargin}
\noindent\textbf{Image inbetweening with inverted latent variables.}
We show another adaptation of outpainting with model inversion by setting two sets of inverted latent variables at two different spatial locations, then perform spatial style fusion between the variables. 
Please refer to Appendix K for implementation details.
As shown in Figure~\ref{fig:inbetween}, we can naturally inbetween~\citep{lu2021cvpr} the area between two images with arbitrary distance.
% and multi-modality.
%
A cyclic panorama of arbitrary width can also be naturally generated by setting the same image on two sides.

%\vspace{\paramargin}
\noindent\textbf{Parallel batching.}
The nature of spatial-independent generation enables parallel inference on a single image.
As shown in Table~\ref{tab:exp-speed_bench}, by stacking a batch of patches together, InfinityGAN can significantly speed up inference at testing up to $7.20$ times.
Note that this speed-up is critical for high-resolution image synthesis with a large number of FLOPs.

%
% \vspace{-1.2em}
%------------------------------------------------------------------------ecrets

%% file: tex/4-tab_quant.tex
% table
\begin{table*}[t!]
\caption{\textbf{Quantitative evaluation on Flickr-Landscape.}
% with ScaleInv FID and user study.} 
Despite we use a disadvantageous setting for our InfinityGAN (discussed in Section~\ref{exp:quantitative}), it still outperforms all baselines after extending the size to 4$\times$ larger. Furthermore, the user study shows an over 90\% preference favors our InfinityGAN results.
The preference is marked as $x$\% when $x$\% of selections prefer the results from the corresponding method over InfinityGAN.
$^\dagger$The images are first resized to 128 before resizing to 197.}
\vspace{-3mm}
\centering
\renewcommand{\arraystretch}{0.8}
\setlength{\tabcolsep}{3.2pt}
\scriptsize
\begin{tabular}{lccccccccccc}
\toprule
% Title
% \multirowcell{2}[-0.6ex][l]{Method} &
% \multicolumn{3}{c}{\multirowcell{2}{Image Size}} &
% \multicolumn{1}{c}{\multirowcell{2}{FID}} &
% \multicolumn{5}{c}{\multirowcell{2}{ScaleInv FID}} &
% \multicolumn{2}{c}{\multirowcell{2}[0em][c]{Preference \\ (v.s. InfinityGAN)}} \\
\multirowcell{2}[-0.6ex][l]{Method} &
\multicolumn{3}{c}{{Image Size}} &
\multicolumn{1}{c}{{FID}} &
\multicolumn{4}{c}{{ScaleInv FID}} &
\multicolumn{2}{c}{Preference  v.s. Ours} &
\multirowcell{2}[-0.6ex][c]{Inference \\ Memory} \\ 
%
%&&&&&&&&&\\
%
% 16x: $342.87$, $389.12$, $320.02$, OOM, $251.45$
%
\cmidrule(lr){2-4}
\cmidrule(lr){5-5}
\cmidrule(lr){6-9}
\cmidrule(lr){10-11}
 &
 Full &
 Train &
 Test 8$\times$& 
 Train &
 1$\times$ & 2$\times$ & 4$\times$ & 8$\times$ & 
 4$\times$ & 8$\times$ & \\ 
\midrule
SinGAN & $128$ & $128$ & $1024$ &
$4.21$ & $4.21$ & $57.10$ & $145.12$ & $210.22$ & 0.80\% & 1.60\% &
$\mathcal{O}(\text{size}^2)$ \\ % Wait for data
COCO-GAN & $128$ & $32$ & $1024$ & % Running on S2
$17.52$ & $41.32$ & $258.51$ & $376.69$ & $387.15$ & 0\% & 0\% &
$\mathcal{O}(1)$ \\
StyleGAN2+NCI & $128$ & $128$ & $1024$ &
$4.19$ & $\textbf{\underline{4.19}}$ & $\textbf{\underline{18.31}}$ & $79.83$ & $189.65$ & 9.20\% & 7.20\% &
$\mathcal{O}(\text{size}^2)$ \\
StyleGAN2+NCI (Patched) & $128$ & $64$ & $1024$ &
$5.35$ & $21.06$ & $58.84$ & $165.65$ & $234.19$ & - & - &
$\mathcal{O}(\text{size}^2)$ \\
StyleGAN2+NCI+PFG & $197^\dagger$ & $101$ & $1576$ & % Running on S5
$86.76$ & $90.79$ & $126.88$ & $211.22$ & $272.80$ & 0.40\% & 1.20\% &
$\mathcal{O}(1)$ \\ 
\midrule
\makecell[l]{ InfinityGAN (Ours) \\ {\scriptsize (StyleGAN2+NCI+PFG+$\structuresyn$)}} & $197^\dagger$ & $101$ & $1576$ &
$11.03$ & $21.84$ & $28.83$ & $\textbf{\underline{61.41}}$ & $\textbf{\underline{121.18}}$ & - & - &
$\mathcal{O}(1)$ \\ \bottomrule
\end{tabular}
\vspace{\tabmargin}
\vspace{-0.5mm}
\label{tab:exp-quantitative}
\end{table*}

%% file: tex/4-fig_quali.tex
\begin{figure}[t!]
    \centering
    \setlength{\tabcolsep}{1.5pt}
    \begin{tabular}{c|c|c|c|c|c}
        \toprule
        & \scriptsize COCO-GAN & \scriptsize SinGAN & \scriptsize StyleGAN+NCI  & \scriptsize	 StyleGAN+NCI+PFG & \scriptsize InfinityGAN (ours) \\ 
        % & & Real (InfinityGAN) \\
        \midrule
        \\ [-1.4em]
        & \small 128$\times$128 & \small 128$\times$128 & \small 128$\times$128 & \small 197$\times$197 & \small 197$\times$197 \\ [-0.2em]
        \hspace{-0.6em}
        \parbox[c]{1.6em}{\rotatebox[origin=c]{90}{\small\makecell{Generated \\[-0.2em] Full Image}\hspace{-4em}}} &
        \includegraphics[width=.182\linewidth]{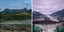} & 
        \includegraphics[width=.182\linewidth]{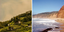} &
        \includegraphics[width=.182\linewidth]{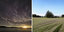} &
        \includegraphics[width=.182\linewidth]{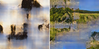} &
        \includegraphics[width=.182\linewidth]{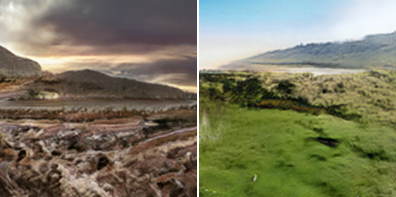} \\ [-0.3em]
        % \parbox[c]{1em}{\rotatebox[origin=c]{90}{Generated \, Full \, Image\hspace{-9.5em}}} &
        % \includegraphics[width=.19\linewidth]{img/quant-plain-gen/cocogan/f_comp.png} & 
        % \includegraphics[width=.19\linewidth]{img/quant-plain-gen/singan/f_comp.png} &
        % \includegraphics[width=.19\linewidth]{img/quant-plain-gen/stylegan2nci/f_comp.png} &
        % \includegraphics[width=.19\linewidth]{img/quant-plain-gen/stylegan2ncifcg/f_comp.png} &
        % \includegraphics[width=.19\linewidth]{img/quant-plain-gen/ours197/f_comp.png} \\ [-0.3em]
        % \parbox[c]{1.3em}{\rotatebox[origin=c]{90}{\raisebox{-9.5pt}{101$\times$101\hspace{-10em}}}} &
        % \includegraphics[width=.19\linewidth]{img/quant-plain-gen/real/patch_sq.png} \\
        %
        \midrule
        \parbox[t]{1.6em}{\rotatebox[origin=c]{90}{\small \makecell{Test-time extension \\[-0.2em] to 1024$\times$1024}\hspace{-7.5em}}} &
        \includegraphics[width=.182\linewidth]{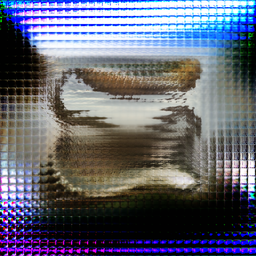} & 
        \includegraphics[width=.182\linewidth]{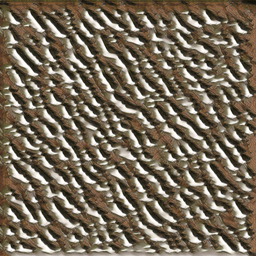} &
        \includegraphics[width=.182\linewidth]{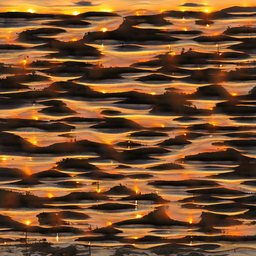} &
        \includegraphics[width=.182\linewidth]{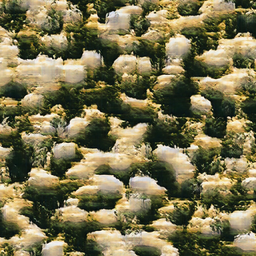} &
        \includegraphics[width=.182\linewidth]{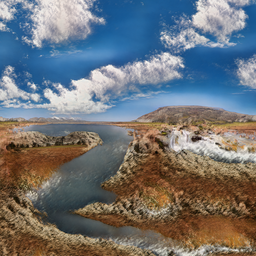} \\ [-0.3em]
        % \multirowcell{2}{\parbox[c]{1.3em}{\rotatebox[origin=c]{90}{\raisebox{-9.5pt}{197$\times$197\hspace{-1em}}}}} &
        % \includegraphics[width=.19\linewidth]{img/quant-plain-gen/real/full_sq_4.png} \\
        %
        %
        % % BACK UP ROW
        % &
        % \includegraphics[width=.24\linewidth]{img/quant-plain-gen/cocogan/1024b.png} & 
        % \includegraphics[width=.24\linewidth]{img/quant-plain-gen/singan/1024a.png} &
        % \includegraphics[width=.24\linewidth]{img/quant-plain-gen/stylegan2nci/1024a.png} &
        % \includegraphics[width=.24\linewidth]{img/quant-plain-gen/ours1024/000007.png} \\
        %
        %
        %
        % &
        % \includegraphics[width=.19\linewidth]{img/quant-plain-gen/real/full_sq_42.png} \\
        % 
        % InfintiyGAN candidates: 7, 11, 25, 27, 30, 31, 32
        \bottomrule
    \end{tabular}
    \vspace{-2mm}
    \caption{
    \textbf{Qualitative comparison.}
    We show that InfinityGAN can produce more favorable holistic appearances against related methods while testing with an extended size 1024$\times$1024. (NCI: Non-Constant Input, PFG: Padding-Free Generator). More results are shown in Appendix E.
    }
    \vspace \figmargin
    \label{fig:exp-qualitative}
    \vspace{-0.5em}
\end{figure}

%% file: tex/4-fig_lsun_and_diversity.tex
% \begin{figure}[t!]
\begin{wrapfigure}[27]{r}{0.38\linewidth}
    \vspace{-1em}
    \centering
    \setlength{\tabcolsep}{1pt}
    \begin{tabular}{cc}
        \includegraphics[width=0.475\linewidth]{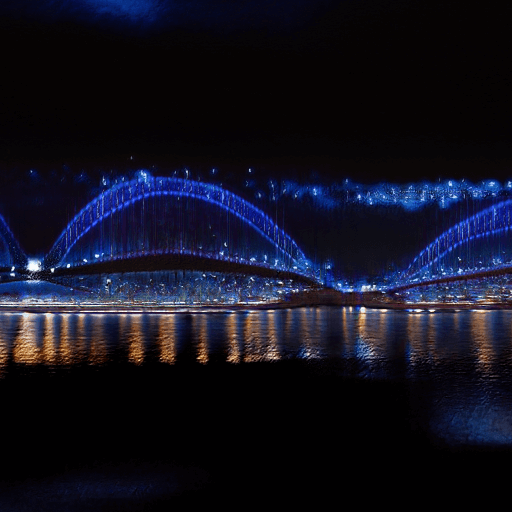} \hfill & \hfill \includegraphics[width=0.475\linewidth]{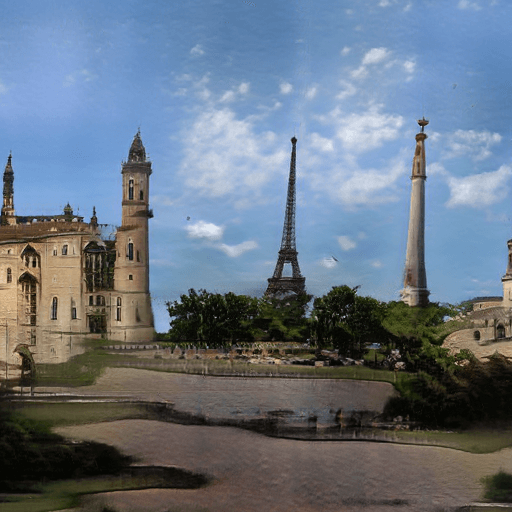}
    \end{tabular}
    \vspace{-1em}
    \caption{
    \textbf{LSUN bridge and tower.}
    InfinityGAN synthesize at 512$\times$512 pixels. We provide more details and samples in Appendix H.
    % We train InfinityGAN with 101$\times$101 pixels patches cropped from 197$\times$197 pixels images, then synthesize at 512$\times$512 pixels. We provide more samples in Appendix.
    }
    \label{fig:lsun}
    % \vspace \figmargin
    % \vspace{1mm}
    \vspace{1em}

    \centering
    \includegraphics[width=\linewidth]{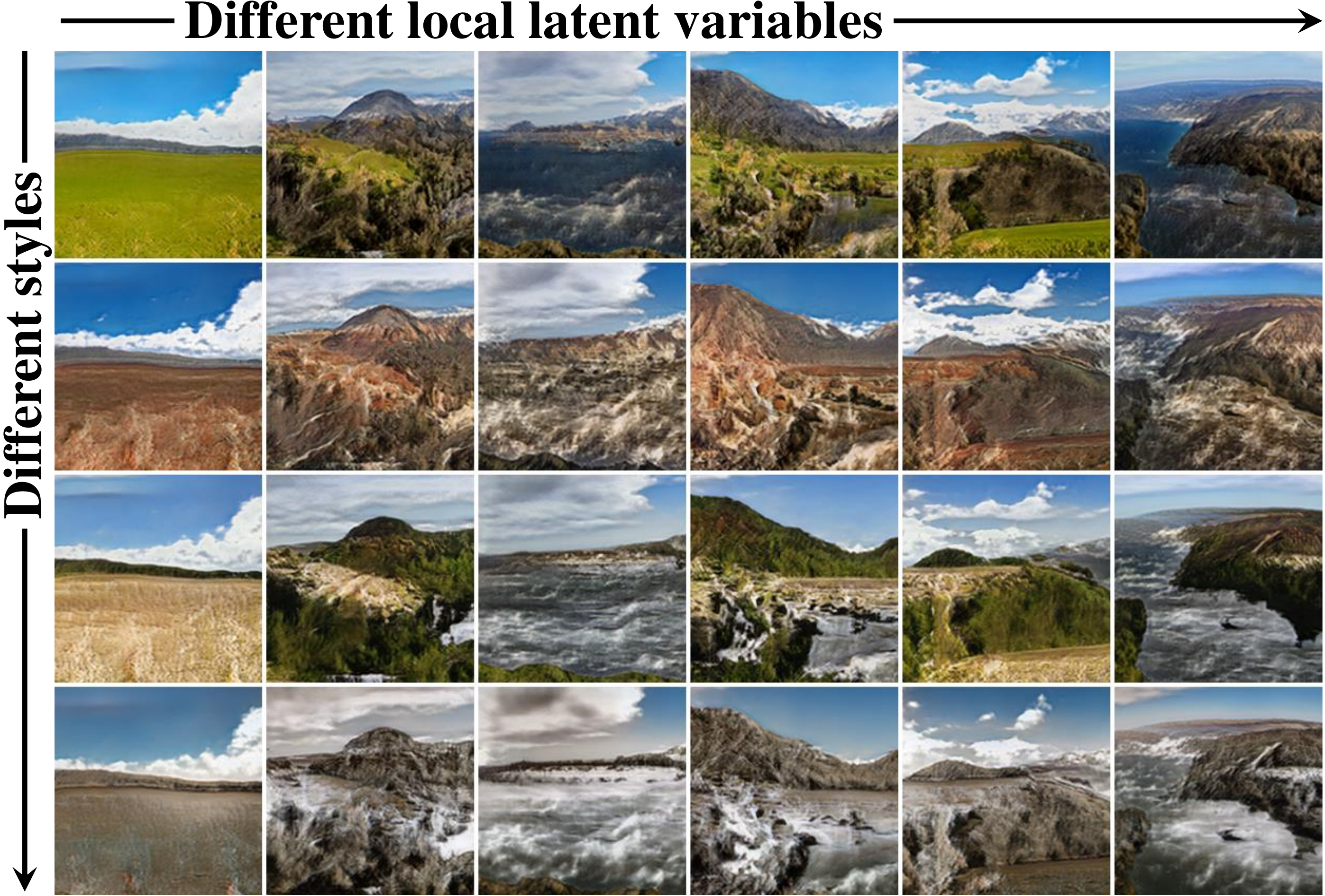}
    \vspace{-2em}
    \caption{
    \textbf{Diversity.}
    InfinityGAN synthesizes diverse samples at the same coordinate with different local latent and styles. More samples are shown in Appendix I.
    % We show that the structure synthesizer and texture synthesizer separately models structure and texture by changing either the local latent $\localz$ or textural latent $\texturez$ while all other variables are fixed.
    % %
    % The results also show that InfinityGAN can synthesize a diverse set of landscape structures at the same coordinate.
    }
    \label{fig:diversity}
    % \vspace \figmargin
    % \vspace{1mm}
% \end{figure}
\end{wrapfigure}

%% file: tex/4-fig_spatial_fusion.tex
\begin{figure*}[t!]
    \centering
    \includegraphics[width=\linewidth]{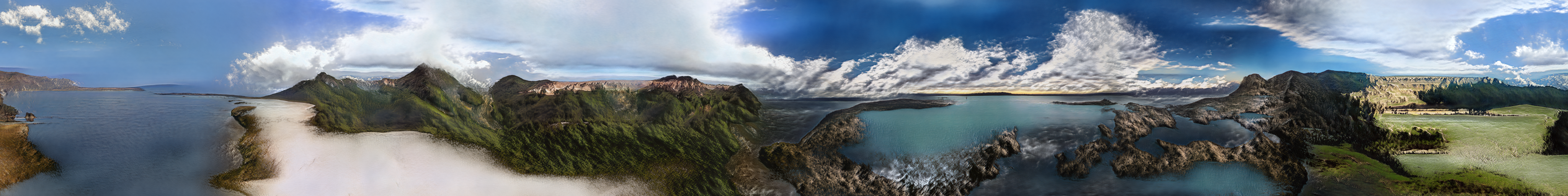}
    \vspace{-6mm}
    \caption{
    \textbf{Spatial style fusion.}
    We present a mechanism in fusing multiple styles together to increase the interestingness and interactiveness of the generation results.
    The 512$\times$4096 image fuses four styles across 258 independently generated patches.
    }
    \vspace \figmargin
    % \vspace{-1em}
    
    \label{fig:spatial_fusion}
\end{figure*}

%% file: tex/4-tab_outpaint_quant_and_speed_bench.tex
% table
\begin{table}[t]
\begin{minipage}{.49\textwidth}
\caption{
\textbf{Outpainting performance.}
The combination of \inout~\citep{inout} and InfinityGAN achieves state-of-the-art IS (higher better) and FID (lower better) performance on image outpainting task.
%We evaluate FID~\citep{fid} (lower better) on two datasets showing a new state-of-the-art performance on image outpainting task by combining \inout~\citep{inout} and our InfinityGAN.
}
\vspace{\tabmargin}
\centering
\scriptsize
\begin{tabular}{lcccc} 
    \toprule
    \multirowcell{2}[-0.6ex][l]{Method} & \multicolumn{2}{c}{Place365} & \multicolumn{2}{c}{Flickr-Scenery} \\
    \cmidrule(lr){2-3}
    \cmidrule(lr){4-5}
    & FID $\shortdownarrow$ & IS $\shortuparrow$ & FID $\shortdownarrow$ & IS $\shortuparrow$ \\
    \midrule
    Boundless  & $35.02$ & $6.15$ & $61.98$ & $6.98$ \\
    NS-outpaint & $50.68$ & $4.70$ & $61.16$ & $4.76$ \\
    % DeepFillv2~\citep{yu2018free}   & $56.14$ & $5.69$ & $62.47$ & $5.38$ \\
    % Image2StyleGAN++~\citep{abdal2020image2styleganpp} & $25.36$ & $6.71$ & $40.39$ & $7.10$ \\
    \midrule
    \inout     & $23.57$ & \underline{\textbf{7.18}} & $30.34$ & $7.16$ \\
    % 
    % % Before
    % \inout{}+ InfinityGAN & \underline{\textbf{11.61}} & \underline{\textbf{16.87}} \\
    % 
    % % After (Current, fixed border)
    \inout{}+ InfinityGAN & \underline{\textbf{9.11}} & 6.78 & \underline{\textbf{15.31}} & \underline{\textbf{7.19}} \\
    % 
    % % No DA (Only places)
    % FID: 14.51
    % IS: 9.47
    \bottomrule
\end{tabular}
\vspace{\tabmargin}
\label{tab:exp-outpainting-quant}
\end{minipage}\hfill
\begin{minipage}{.49\textwidth}
\caption{\textbf{Inference speed up with parallel batching.} 
Benefit from the spatial independent generation nature,  InfinityGAN achieves up to 7.20$\times$ inference speed up by with parallel batching at 8192$\times$8192 pixels.
%We show that InfinityGAN can achieve up to 7.20$\times$ inference speed up by with parallel batching, which is achieved by leveraging the spatial independent generation nature of InfinityGAN.
%
The complete table can be found in Appendix P.}
\vspace{-3mm}
\centering
\setlength{\tabcolsep}{3.5pt}
% Please add the following required packages to your document preamble:
% \usepackage{booktabs}
% \usepackage{multirow}
\scriptsize
\begin{tabular}{lcccc}
\toprule
% Title
\makecell[l]{Method} &
\makecell{Parallel \\ Batch Size} &
\makecell{\# GPUs} &
\makecell{Inference Time \\ (second / image)} &
Speed Up \\ 
\midrule
StyleGAN2
& N/A & $1$ & 
OOM & - \\
\midrule
\multirowcell{2}[-0.6ex][l]{Ours}
& $1$ & $1$ &
$137.44$ &
$\times 1.00$ \\
%
% \cmidrule(lr){3-9}
%
& $128$ & $8$ &
$19.09$ &
$\times 7.20$ \\
\bottomrule
\end{tabular}
\vspace{\tabmargin}
\label{tab:exp-speed_bench}
\end{minipage}
\vspace{-.5em}
\end{table}

%% Version with IS

% \begin{tabular}{lcccc} 
%     \toprule
%     \multirowcell{2}[-0.6ex][l]{Method} & \multicolumn{2}{c}{Place365} & \multicolumn{2}{c}{Flickr} \\
%     \cmidrule(r){2-3} \cmidrule(r){4-5}
%      & {FID $\shortdownarrow$ } & {IS $\shortuparrow$ }  & {FID $\shortdownarrow$ } &{IS $\shortuparrow$ }  \\
%     \midrule
%     Boundless~\citep{teterwak2019boundless}  & $35.02$ & $6.15$ & $61.98$ & $6.98$ \\
%     NS-outpaint~\citep{yang2019very} & $50.68$ & $4.70$ & $61.16$ & $4.76$ \\
%     DeepFillv2~\citep{yu2018free}   & $56.14$ & $5.69$ & $62.47$ & $5.38$ \\
%     mage2StyleGAN++~\citep{abdal2020image2styleganpp} & $25.36$ & $6.71$ & $40.39$ & $7.10$ \\
%     \midrule
%     MethodX     & $23.57$ & $7.18$ & $30.34$ & $7.16$ \\
%     MethodX-C   & $29.24$ & $7.69$ & $33.17$ & $7.15$ \\
%     \midrule
%     MethodX + InfinityGAN & $11.61$ & $6.49$  & $16.87$ & $6.68$ \\
%     \bottomrule
% \end{tabular}
% \vspace{\tabmargin}
% \label{tab:exp-outpainting-quant}
% \end{table}

%% file: tex/4-fig_outpaint_quali.tex
% Input => Boundless, DeepFillV2, NS-outpaint, InOut, ours
% Panorama
% 2 sets, double-column

\begin{figure}[t]
\begin{minipage}{.49\textwidth}
    \centering
    \includegraphics[width=\linewidth]{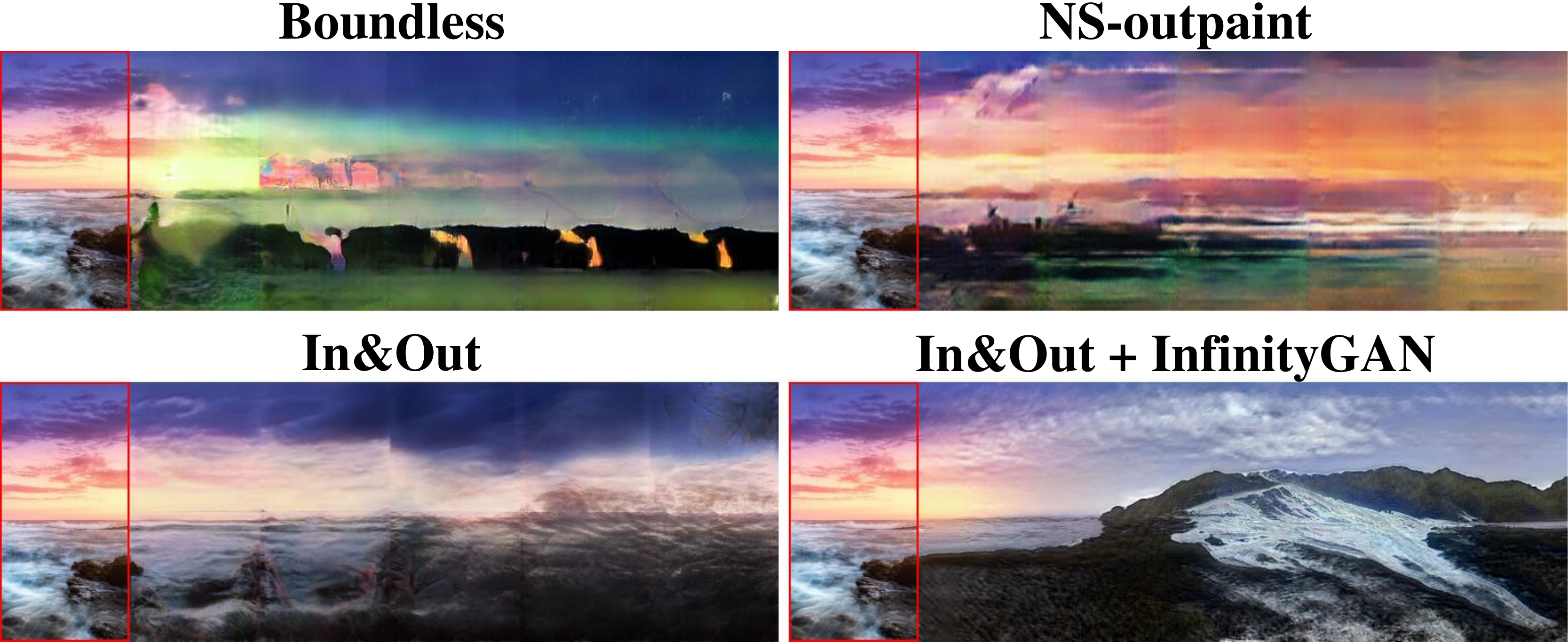}
    \vspace{-6mm}
    \caption{
    \textbf{Outpainting long-range area.}
    InfinityGAN synthesizes continuous and more plausible outpainting results for arbitrarily large outpainting areas. The real image annotated with {\color{red}red} box is 256$\times$128 pixels.
    }
    \vspace \figmargin
    % \vspace{-2mm}
    \label{fig:outpaint-quali}
\end{minipage} \hfill
\begin{minipage}{.49\textwidth}
    \centering
    \includegraphics[width=\linewidth]{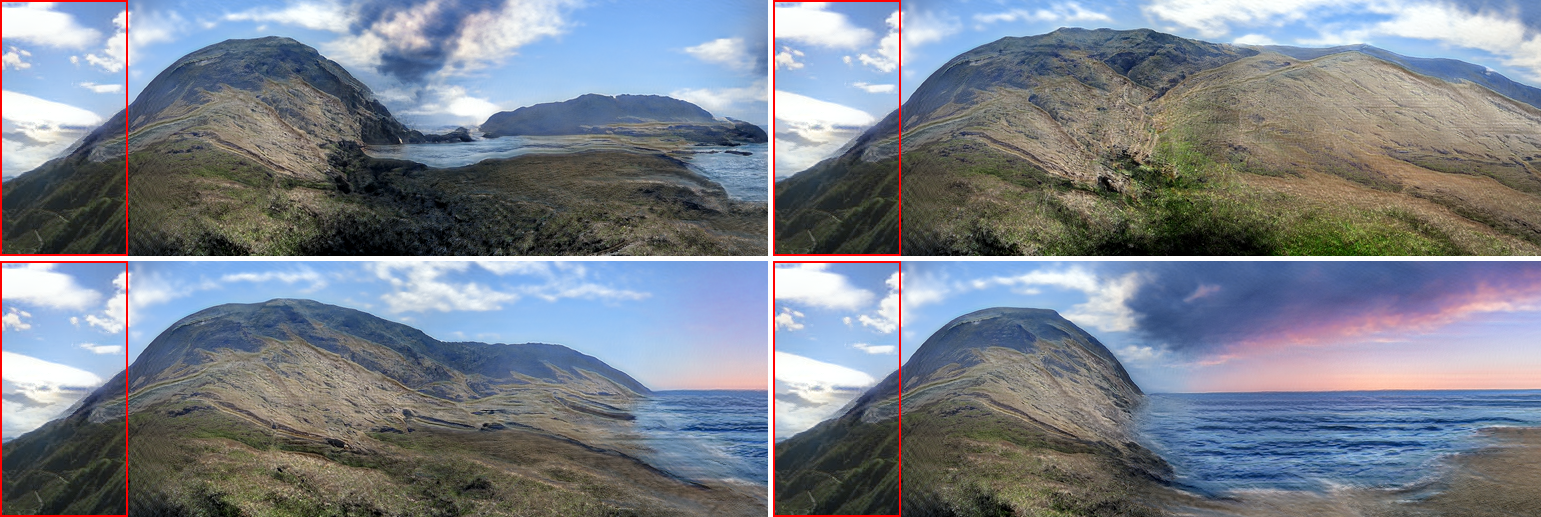}
    \vspace{-6mm}
    \caption{
    \textbf{Multi-modal outpainting.} InfinityGAN can natively achieve multi-modal outpainting by sampling different local latents in the outpainted region.
    The real image annotated with {\color{red}red} box is 256$\times$128 pixels.
    We present more outpainting samples in Appendix M.
    }
    \vspace \figmargin
    \vspace{1mm}
    \label{fig:outpaint-multimodal}
\end{minipage}
\end{figure}

%% file: tex/4-fig_inbetweening.tex
\begin{figure*}[t]
    \centering
    \includegraphics[width=\linewidth]{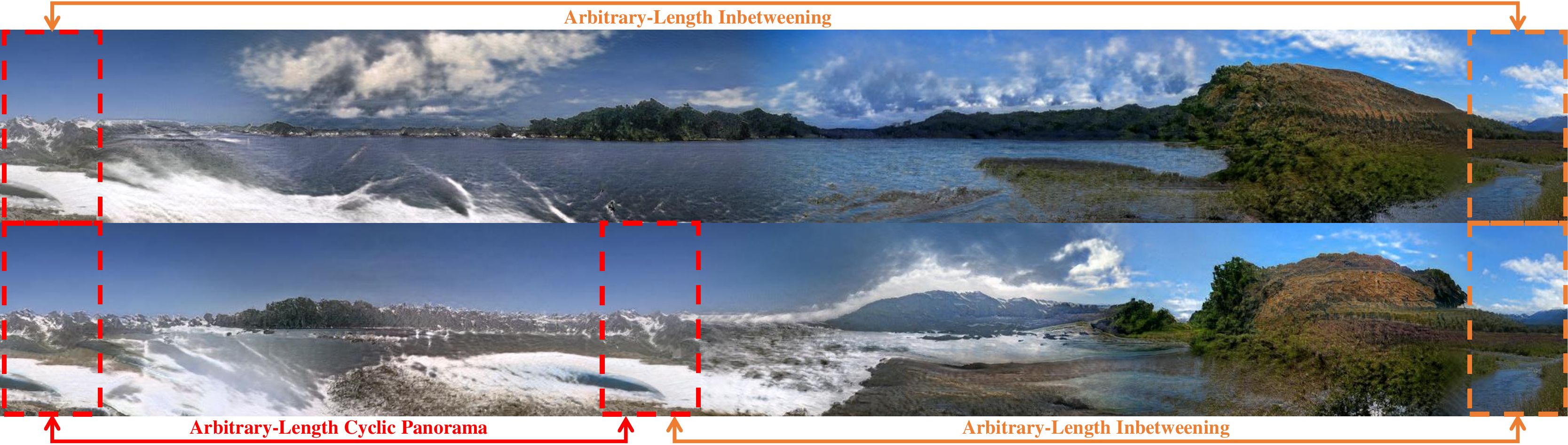}
    \vspace{-7mm}
    \caption{
    \textbf{Image inbetweening with inverted latents.} 
    %
    %\hubert{Resolving critical issue in panorama generation.}
    The InfinityGAN can synthesize arbitrary-length cyclic panorama and inbetweened images by inverting a real image at different position.
    The top-row image size is 256$\times$2080 pixels.
    We present more samples in Appendix N and Appendix I.
    }
    \vspace \figmargin
    \label{fig:inbetween}
\end{figure*}

%% file: 5-conclusion.tex
\vspace{\secmargin}
\section{Conclusions}
\vspace{\secmargin}
In this work, we propose and tackle the problem of synthesizing infinite-pixel images, and demonstrate several applications of  InfinityGAN, including image outpainting and inbetweening. 

%There is still room for improvement. 
Our future work will focus on improving InfinityGAN in several aspects.
First, our Flickr-Landscape dataset consists of images taken at different FoVs and distances to the scenes. 
When InfinityGAN composes landscapes of different scales together, synthesized images may contain artifacts. 
Second, similar to the FoV problem, some images intentionally include tree leaves on top of the image as a part of the photography composition. 
These greenish textures cause InfinityGAN sometimes synthesizing trees or related elements in the sky region.
%MH: why do you say "bias"? why it leads to "floating islands in the sky"? If you cannot say it clearly, then you do not need to say it. 
%Such bias can mislead InfinityGAN to synthesize floating islands in the sky.
% Such large textured regions can cause InfinityGAN to synthesize inconsistent images. 
%
Third, there is still a slight decrease in FID score in comparison to StyleGAN2. 
This may be related to the convergence problem in video synthesis~\citep{tian2021video}, in which the generator achieves inferior performance if a preceding network (\eg the motion module in video synthesis) is jointly trained with the image module.

%% file: 7-ethics-statement.tex
\section{Ethics Statement}
Our work follows the General Ethical Principles listed at ICLR Code of Ethics (\url{https://iclr.cc/public/CodeOfEthics}). 

The research in generative modeling is frequently accompanied by concerns about the misuse of manipulating or hallucinating information for improper use. Despite none of the proposed techniques aiming at improving manipulation of fine-grained image detail or hallucinating human activities, we cannot rule out the potential of misusing the framework to recreate fake scenery images for any inappropriate application. However, as we do not drastically alter the plausibility of synthesis results in the high-frequency domain, our research is still covered by continuing research in ethical generative modeling and image forensics.

%% file: main.bbl
\begin{thebibliography}{52}
\providecommand{\natexlab}[1]{#1}
\providecommand{\url}[1]{\texttt{#1}}
\expandafter\ifx\csname urlstyle\endcsname\relax
  \providecommand{\doi}[1]{doi: #1}\else
  \providecommand{\doi}{doi: \begingroup \urlstyle{rm}\Url}\fi

\bibitem[Abdal et~al.(2020)Abdal, Qin, and Wonka]{abdal2020image2styleganpp}
Rameen Abdal, Yipeng Qin, and Peter Wonka.
\newblock Image2stylegan++: How to edit the embedded images?
\newblock In \emph{IEEE Conference on Computer Vision and Pattern Recognition},
  2020.

\bibitem[Balaji et~al.(2021)Balaji, Sajedi, Kalibhat, Ding, St{\"o}ger,
  Soltanolkotabi, and Feizi]{balaji2021understanding}
Yogesh Balaji, Mohammadmahdi Sajedi, Neha~Mukund Kalibhat, Mucong Ding, Dominik
  St{\"o}ger, Mahdi Soltanolkotabi, and Soheil Feizi.
\newblock Understanding over-parameterization in generative adversarial
  networks.
\newblock In \emph{International Conference on Learning Representations}, 2021.

\bibitem[Bergmann et~al.(2017)Bergmann, Jetchev, and
  Vollgraf]{bergmann2017learning}
Urs Bergmann, Nikolay Jetchev, and Roland Vollgraf.
\newblock Learning texture manifolds with the periodic spatial gan.
\newblock 2017.

\bibitem[Chan et~al.(2021)Chan, Monteiro, Kellnhofer, Wu, and
  Wetzstein]{chan2021pi}
Eric~R Chan, Marco Monteiro, Petr Kellnhofer, Jiajun Wu, and Gordon Wetzstein.
\newblock pi-gan: Periodic implicit generative adversarial networks for
  3d-aware image synthesis.
\newblock In \emph{IEEE Conference on Computer Vision and Pattern Recognition},
  2021.

\bibitem[Chen et~al.(2021)Chen, Liu, and Wang]{chen2020liif}
Yinbo Chen, Sifei Liu, and Xiaolong Wang.
\newblock Learning continuous image representation with local implicit image
  function.
\newblock In \emph{IEEE Conference on Computer Vision and Pattern Recognition},
  2021.

\bibitem[Cheng et~al.(2021)Cheng, Lin, Lee, Tulyakov, Ren, and Yang]{inout}
Yen-Chi Cheng, Chieh~Hubert Lin, Hsin-Ying Lee, Sergey Tulyakov, Jian Ren, and
  Ming-Hsuan Yang.
\newblock In\&{O}ut: Diverse image outpainting via gan inversion.
\newblock \emph{arXiv preprint arXiv:2104.00675}, 2021.

\bibitem[DeVries et~al.(2021)DeVries, Bautista, Srivastava, Taylor, and
  Susskind]{devries2021unconstrained}
Terrance DeVries, Miguel~Angel Bautista, Nitish Srivastava, Graham~W Taylor,
  and Joshua~M Susskind.
\newblock Unconstrained scene generation with locally conditioned radiance
  fields.
\newblock In \emph{IEEE International Conference on Computer Vision}, 2021.

\bibitem[Efros \& Leung(1999)Efros and Leung]{efros1999texture}
Alexei~A Efros and Thomas~K Leung.
\newblock Texture synthesis by non-parametric sampling.
\newblock In \emph{IEEE International Conference on Computer Vision}, 1999.

\bibitem[Esser et~al.(2021)Esser, Rombach, and Ommer]{esser2021taming}
Patrick Esser, Robin Rombach, and Bjorn Ommer.
\newblock Taming transformers for high-resolution image synthesis.
\newblock In \emph{IEEE Conference on Computer Vision and Pattern Recognition},
  2021.

\bibitem[Fr{\"u}hst{\"u}ck et~al.(2019)Fr{\"u}hst{\"u}ck, Alhashim, and
  Wonka]{fruhstuck2019tilegan}
Anna Fr{\"u}hst{\"u}ck, Ibraheem Alhashim, and Peter Wonka.
\newblock Tilegan: synthesis of large-scale non-homogeneous textures.
\newblock \emph{ACM Transactions on Graphics}, 2019.

\bibitem[Goodfellow et~al.(2014)Goodfellow, Pouget-Abadie, Mirza, Xu,
  Warde-Farley, Ozair, Courville, and Bengio]{goodfellow2014generative}
Ian Goodfellow, Jean Pouget-Abadie, Mehdi Mirza, Bing Xu, David Warde-Farley,
  Sherjil Ozair, Aaron Courville, and Yoshua Bengio.
\newblock Generative adversarial nets.
\newblock In \emph{Neural Information Processing Systems}, 2014.

\bibitem[Gulrajani et~al.(2017)Gulrajani, Ahmed, Arjovsky, Dumoulin, and
  Courville]{wgangp}
Ishaan Gulrajani, Faruk Ahmed, Martín Arjovsky, Vincent Dumoulin, and Aaron~C.
  Courville.
\newblock Improved training of wasserstein gans.
\newblock In \emph{Neural Information Processing Systems}, 2017.

\bibitem[Heusel et~al.(2017)Heusel, Ramsauer, Unterthiner, Nessler, and
  Hochreiter]{fid}
Martin Heusel, Hubert Ramsauer, Thomas Unterthiner, Bernhard Nessler, and Sepp
  Hochreiter.
\newblock {GANs} trained by a two time-scale update rule converge to a local
  nash equilibrium.
\newblock In \emph{Neural Information Processing Systems}, 2017.

\bibitem[Hinz et~al.(2021)Hinz, Fisher, Wang, and Wermter]{hinz2021improved}
Tobias Hinz, Matthew Fisher, Oliver Wang, and Stefan Wermter.
\newblock Improved techniques for training single-image gans.
\newblock In \emph{IEEE Winter Conference on Applications of Computer Vision},
  2021.

\bibitem[Huang \& Belongie(2017)Huang and Belongie]{huang2017adain}
Xun Huang and Serge Belongie.
\newblock Arbitrary style transfer in real-time with adaptive instance
  normalization.
\newblock In \emph{IEEE International Conference on Computer Vision}, 2017.

\bibitem[Isola et~al.(2017)Isola, Zhu, Zhou, and Efros]{isola2017image}
Phillip Isola, Jun-Yan Zhu, Tinghui Zhou, and Alexei~A Efros.
\newblock Image-to-image translation with conditional adversarial networks.
\newblock In \emph{IEEE Conference on Computer Vision and Pattern Recognition},
  2017.

\bibitem[Jetchev et~al.(2018)Jetchev, Bergmann, and Yildirim]{jetchev2018copy}
Nikolay Jetchev, Urs Bergmann, and Gokhan Yildirim.
\newblock Copy the old or paint anew? an adversarial framework for (non-)
  parametric image stylization.
\newblock \emph{Neural Information Processing Systems Workshops}, 2018.

\bibitem[Karras et~al.(2018)Karras, Aila, Laine, and
  Lehtinen]{karras2018progressive}
Tero Karras, Timo Aila, Samuli Laine, and Jaakko Lehtinen.
\newblock Progressive growing of gans for improved quality, stability, and
  variation.
\newblock In \emph{International Conference on Learning Representations}, 2018.

\bibitem[Karras et~al.(2020)Karras, Laine, Aittala, Hellsten, Lehtinen, and
  Aila]{karras2020analyzing}
Tero Karras, Samuli Laine, Miika Aittala, Janne Hellsten, Jaakko Lehtinen, and
  Timo Aila.
\newblock Analyzing and improving the image quality of stylegan.
\newblock In \emph{IEEE Conference on Computer Vision and Pattern Recognition},
  2020.

\bibitem[Kingma \& Ba(2015)Kingma and Ba]{adam}
Diederik Kingma and Jimmy Ba.
\newblock Adam: A method for stochastic optimization.
\newblock In \emph{International Conference on Learning Representations}, 2015.

\bibitem[Lee et~al.(2020)Lee, Tseng, Mao, Huang, Lu, Singh, and
  Yang]{DRIT_plus}
Hsin-Ying Lee, Hung-Yu Tseng, Qi~Mao, Jia-Bin Huang, Yu-Ding Lu, Maneesh~Kumar
  Singh, and Ming-Hsuan Yang.
\newblock Drit++: Diverse image-to-image translation viadisentangled
  representations.
\newblock \emph{International Journal of Computer Vision}, pp.\  1--16, 2020.

\bibitem[Lin et~al.(2019)Lin, Chang, Chen, Juan, Wei, and Chen]{lin2019cocogan}
Chieh~Hubert Lin, Chia-Che Chang, Yu-Sheng Chen, Da-Cheng Juan, Wei Wei, and
  Hwann-Tzong Chen.
\newblock {COCO-GAN}: Generation by parts via conditional coordinating.
\newblock In \emph{IEEE International Conference on Computer Vision}, 2019.

\bibitem[Liu et~al.(2021)Liu, Tucker, Jampani, Makadia, Snavely, and
  Kanazawa]{liu2020infinite}
Andrew Liu, Richard Tucker, Varun Jampani, Ameesh Makadia, Noah Snavely, and
  Angjoo Kanazawa.
\newblock Infinite nature: Perpetual view generation of natural scenes from a
  single image.
\newblock In \emph{IEEE International Conference on Computer Vision}, 2021.

\bibitem[Liu et~al.(2018{\natexlab{a}})Liu, Reda, Shih, Wang, Tao, and
  Catanzaro]{liu2018image}
Guilin Liu, Fitsum~A Reda, Kevin~J Shih, Ting-Chun Wang, Andrew Tao, and Bryan
  Catanzaro.
\newblock Image inpainting for irregular holes using partial convolutions.
\newblock In \emph{European Conference on Computer Vision}, 2018{\natexlab{a}}.

\bibitem[Liu et~al.(2018{\natexlab{b}})Liu, Lehman, Molino, Petroski~Such,
  Frank, Sergeev, and Yosinski]{liu2018coordconv}
Rosanne Liu, Joel Lehman, Piero Molino, Felipe Petroski~Such, Eric Frank, Alex
  Sergeev, and Jason Yosinski.
\newblock An intriguing failing of convolutional neural networks and the
  coordconv solution.
\newblock In \emph{Neural Information Processing Systems}, 2018{\natexlab{b}}.

\bibitem[Lu et~al.(2021)Lu, Chang, and Chiu]{lu2021cvpr}
Chia-Ni Lu, Ya-Chu Chang, and Wei-Chen Chiu.
\newblock Bridging the visual gap: Wide-range image blending.
\newblock In \emph{IEEE Conference on Computer Vision and Pattern Recognition},
  2021.

\bibitem[Mao et~al.(2019)Mao, Lee, Tseng, Ma, and Yang]{mao2019mode}
Qi~Mao, Hsin-Ying Lee, Hung-Yu Tseng, Siwei Ma, and Ming-Hsuan Yang.
\newblock Mode seeking generative adversarial networks for diverse image
  synthesis.
\newblock In \emph{IEEE Conference on Computer Vision and Pattern Recognition},
  2019.

\bibitem[Mescheder et~al.(2018)Mescheder, Geiger, and
  Nowozin]{mescheder2018r1reg}
Lars Mescheder, Andreas Geiger, and Sebastian Nowozin.
\newblock Which training methods for gans do actually converge?
\newblock In \emph{International Conference on Machine Learning}, 2018.

\bibitem[Mescheder et~al.(2019)Mescheder, Oechsle, Niemeyer, Nowozin, and
  Geiger]{mescheder2019occupancy}
Lars Mescheder, Michael Oechsle, Michael Niemeyer, Sebastian Nowozin, and
  Andreas Geiger.
\newblock Occupancy networks: Learning 3d reconstruction in function space.
\newblock In \emph{IEEE Conference on Computer Vision and Pattern Recognition},
  2019.

\bibitem[Mildenhall et~al.(2020)Mildenhall, Srinivasan, Tancik, Barron,
  Ramamoorthi, and Ng]{mildenhall2020nerf}
Ben Mildenhall, Pratul~P Srinivasan, Matthew Tancik, Jonathan~T Barron, Ravi
  Ramamoorthi, and Ren Ng.
\newblock Nerf: Representing scenes as neural radiance fields for view
  synthesis.
\newblock In \emph{European Conference on Computer Vision}, 2020.

\bibitem[Nazeri et~al.(2019)Nazeri, Ng, Joseph, Qureshi, and
  Ebrahimi]{Nazeri_2019_ICCV}
Kamyar Nazeri, Eric Ng, Tony Joseph, Faisal Qureshi, and Mehran Ebrahimi.
\newblock Edgeconnect: Structure guided image inpainting using edge prediction.
\newblock In \emph{IEEE International Conference on Computer Vision Workshops},
  2019.

\bibitem[Niemeyer \& Geiger(2021)Niemeyer and Geiger]{niemeyer2021giraffe}
Michael Niemeyer and Andreas Geiger.
\newblock Giraffe: Representing scenes as compositional generative neural
  feature fields.
\newblock In \emph{IEEE Conference on Computer Vision and Pattern Recognition},
  2021.

\bibitem[Odena et~al.(2017)Odena, Olah, and Shlens]{odena2017acgan}
Augustus Odena, Christopher Olah, and Jonathon Shlens.
\newblock Conditional image synthesis with auxiliary classifier {GAN}s.
\newblock In \emph{International Conference on Machine Learning}, 2017.

\bibitem[Oord et~al.(2016)Oord, Kalchbrenner, and Kavukcuoglu]{oord2016pixel}
Aaron van~den Oord, Nal Kalchbrenner, and Koray Kavukcuoglu.
\newblock Pixel recurrent neural networks.
\newblock In \emph{International Conference on Machine Learning}, 2016.

\bibitem[Park et~al.(2019)Park, Florence, Straub, Newcombe, and
  Lovegrove]{part2019deepsdf}
Jeong~Joon Park, Peter Florence, Julian Straub, Richard Newcombe, and Steven
  Lovegrove.
\newblock Deepsdf: Learning continuous signed distance functions for shape
  representation.
\newblock In \emph{IEEE Conference on Computer Vision and Pattern Recognition},
  2019.

\bibitem[Razavi et~al.(2019)Razavi, Oord, and Vinyals]{razavi2019vqvae2}
Ali Razavi, Aaron van~den Oord, and Oriol Vinyals.
\newblock Generating diverse high-fidelity images with {VQ-VAE-2}.
\newblock In \emph{Neural Information Processing Systems}, 2019.

\bibitem[Sabini \& Rusak(2018)Sabini and Rusak]{sabini2018outpainting}
Mark Sabini and Gili Rusak.
\newblock Painting outside the box: Image outpainting with gans.
\newblock \emph{arXiv preprint arXiv:1808.08483}, 2018.

\bibitem[Shaham et~al.(2019)Shaham, Dekel, and Michaeli]{shaham2019singan}
Tamar~Rott Shaham, Tali Dekel, and Tomer Michaeli.
\newblock Singan: Learning a generative model from a single natural image.
\newblock In \emph{IEEE International Conference on Computer Vision}, 2019.

\bibitem[Shocher et~al.(2019)Shocher, Bagon, Isola, and
  Irani]{shocher2019ingan}
Assaf Shocher, Shai Bagon, Phillip Isola, and Michal Irani.
\newblock Ingan: Capturing and retargeting the" dna" of a natural image.
\newblock In \emph{IEEE International Conference on Computer Vision}, 2019.

\bibitem[Sitzmann et~al.(2020)Sitzmann, Martel, Bergman, Lindell, and
  Wetzstein]{sitzmann2020implicit}
Vincent Sitzmann, Julien Martel, Alexander Bergman, David Lindell, and Gordon
  Wetzstein.
\newblock Implicit neural representations with periodic activation functions.
\newblock In \emph{Neural Information Processing Systems}, 2020.

\bibitem[Skorokhodov et~al.(2021)Skorokhodov, Sotnikov, and
  Elhoseiny]{skorokhodov2021aligning}
Ivan Skorokhodov, Grigorii Sotnikov, and Mohamed Elhoseiny.
\newblock Aligning latent and image spaces to connect the unconnectable.
\newblock In \emph{IEEE International Conference on Computer Vision}, 2021.

\bibitem[Tancik et~al.(2020)Tancik, Srinivasan, Mildenhall, Fridovich-Keil,
  Raghavan, Singhal, Ramamoorthi, Barron, and Ng]{tancik2020fourier}
Matthew Tancik, Pratul~P Srinivasan, Ben Mildenhall, Sara Fridovich-Keil,
  Nithin Raghavan, Utkarsh Singhal, Ravi Ramamoorthi, Jonathan~T Barron, and
  Ren Ng.
\newblock Fourier features let networks learn high frequency functions in low
  dimensional domains.
\newblock In \emph{Neural Information Processing Systems}, 2020.

\bibitem[Teterwak et~al.(2019)Teterwak, Sarna, Krishnan, Maschinot, Belanger,
  Liu, and Freeman]{teterwak2019boundless}
Piotr Teterwak, Aaron Sarna, Dilip Krishnan, Aaron Maschinot, David Belanger,
  Ce~Liu, and William~T Freeman.
\newblock Boundless: Generative adversarial networks for image extension.
\newblock In \emph{IEEE International Conference on Computer Vision}, 2019.

\bibitem[Tian et~al.(2021)Tian, Ren, Chai, Olszewski, Peng, Metaxas, and
  Tulyakov]{tian2021video}
Yu~Tian, Jian Ren, Menglei Chai, Kyle Olszewski, Xi~Peng, Dimitris~N. Metaxas,
  and Sergey Tulyakov.
\newblock A good image generator is what you need for high-resolution video
  synthesis.
\newblock In \emph{International Conference on Learning Representations}, 2021.

\bibitem[Vaswani et~al.(2017)Vaswani, Shazeer, Parmar, Uszkoreit, Jones, Gomez,
  Kaiser, and Polosukhin]{vaswani2017attention}
Ashish Vaswani, Noam Shazeer, Niki Parmar, Jakob Uszkoreit, Llion Jones,
  Aidan~N Gomez, {\L}ukasz Kaiser, and Illia Polosukhin.
\newblock Attention is all you need.
\newblock In \emph{Neural Information Processing Systems}, 2017.

\bibitem[Wulff \& Torralba(2020)Wulff and Torralba]{wulff2020improving}
Jonas Wulff and Antonio Torralba.
\newblock Improving inversion and generation diversity in stylegan using a
  gaussianized latent space.
\newblock \emph{arXiv preprint arXiv:2009.06529}, 2020.

\bibitem[Xian et~al.(2018)Xian, Sangkloy, Agrawal, Raj, Lu, Fang, Yu, and
  Hays]{xian2018texturegan}
Wenqi Xian, Patsorn Sangkloy, Varun Agrawal, Amit Raj, Jingwan Lu, Chen Fang,
  Fisher Yu, and James Hays.
\newblock Texturegan: Controlling deep image synthesis with texture patches.
\newblock In \emph{IEEE Conference on Computer Vision and Pattern Recognition},
  2018.

\bibitem[Xu et~al.(2021)Xu, Wang, Chen, Zhou, and Loy]{xu2021positional}
Rui Xu, Xintao Wang, Kai Chen, Bolei Zhou, and Chen~Change Loy.
\newblock Positional encoding as spatial inductive bias in gans.
\newblock In \emph{IEEE Conference on Computer Vision and Pattern Recognition},
  2021.

\bibitem[Yang et~al.(2019)Yang, Dong, Liu, Yang, and Yan]{yang2019very}
Zongxin Yang, Jian Dong, Ping Liu, Yi~Yang, and Shuicheng Yan.
\newblock Very long natural scenery image prediction by outpainting.
\newblock In \emph{IEEE International Conference on Computer Vision}, 2019.

\bibitem[Yu et~al.(2019)Yu, Lin, Yang, Shen, Lu, and Huang]{yu2018free}
Jiahui Yu, Zhe Lin, Jimei Yang, Xiaohui Shen, Xin Lu, and Thomas~S Huang.
\newblock Free-form image inpainting with gated convolution.
\newblock In \emph{IEEE International Conference on Computer Vision}, 2019.

\bibitem[Zhang et~al.(2018)Zhang, Isola, Efros, Shechtman, and
  Wang]{zhang2018perceptual}
Richard Zhang, Phillip Isola, Alexei~A Efros, Eli Shechtman, and Oliver Wang.
\newblock The unreasonable effectiveness of deep features as a perceptual
  metric.
\newblock In \emph{IEEE Conference on Computer Vision and Pattern Recognition},
  2018.

\bibitem[Zhou et~al.(2017)Zhou, Lapedriza, Khosla, Oliva, and
  Torralba]{zhou2017places}
Bolei Zhou, Agata Lapedriza, Aditya Khosla, Aude Oliva, and Antonio Torralba.
\newblock Places: A 10 million image database for scene recognition.
\newblock \emph{IEEE Transactions on Pattern Analysis and Machine
  Intelligence}, 2017.

\end{thebibliography}
